\documentclass{article}
\usepackage[utf8]{inputenc}
\usepackage[a4paper, total={6in, 8in}]{geometry}
\usepackage{hyperref}
\usepackage{authblk}
\usepackage{apacite}
\usepackage{graphicx}
\usepackage{caption}
\usepackage{natbib}

\newcounter{examplenumi}
\newenvironment{examplenumerate}{\begin{enumerate} \itshape \setcounter{enumi}{\theexamplenumi}}{ \setcounter{examplenumi}{\theenumi}\end{enumerate}}

\usepackage{pifont}
\newcommand{\cmark}{\ding{51}}%
\newcommand{\xmark}{\ding{55}}

\title{Automatic Detection of Cyberbullying in Social Media Text}

\author[1]{Cynthia Van Hee}
\author[1]{Gilles Jacobs}
\author[2]{Chris Emmery}
\author[1]{Bart Desmet}
\author[1]{Els Lefever}
\author[2]{Ben Verhoeven}
\author[2]{Guy De Pauw}
\author[2]{Walter Daelemans}
\author[1]{V\'{e}ronique Hoste}
\affil[1]{LT3, Ghent University}
\affil[2]{CLiPS, University of Antwerp}
\affil[1]{\textit {\{cynthia.vanhee, gillesm.jacobs, bart.desmet, els.lefever, veronique.hoste\}@ugent.be}}
\affil[2]{\textit {\{ben.verhoeven, guy.depauw, walter.daelemans\}@uantwerpen.be}}
\affil[2]{\textit{c.d.emmery@uvt.nl}}
\date{}

\begin{document}

\maketitle

\begin{abstract}
While social media offer great communication opportunities, they also increase the vulnerability of young people to threatening situations online. Recent studies report that cyberbullying constitutes a growing problem among youngsters. Successful prevention depends on the adequate detection of potentially harmful messages and the information overload on the Web requires intelligent systems to identify potential risks automatically. The focus of this paper is on automatic cyberbullying detection in social media text by modelling posts written by bullies, victims, and bystanders of online bullying. We describe the collection and fine-grained annotation of a training corpus for English and Dutch and perform a series of binary classification experiments to determine the feasibility of automatic cyberbullying detection. We make use of linear support vector machines exploiting a rich feature set and investigate which information sources contribute the most for this particular task. Experiments on a holdout test set reveal promising results for the detection of cyberbullying-related posts. After optimisation of the hyperparameters, the classifier yields an F$_1$-score of 64\% and 61\% for English and Dutch respectively, and considerably outperforms baseline systems based on keywords and word unigrams.
\end{abstract}

\section{Introduction}
\label{intro}
Web 2.0 has had a substantial impact on communication and relationships in today's society. Children and teenagers go online more frequently, at younger ages, and in more diverse ways (e.g.~smartphones, laptops and tablets). Although most of teenagers' Internet use is harmless and the benefits of digital communication are evident, the freedom and anonymity experienced online makes young people vulnerable, with cyberbullying being one of the major threats~\citep{Livingstone2010,Tokunaga2010,Livingstone2013}.

Bullying is not a new phenomenon, and cyberbullying has manifested itself as soon as digital technologies have become primary communication tools. On the positive side, social media like blogs, social networking sites (e.g.~Facebook) and instant messaging platforms (e.g.~WhatsApp) make it possible to communicate with anyone and at any time. Moreover, they are a place where people engage in social interaction, offering the possibility to establish new relationships and maintain existing friendships~\citep{Gross2002,Mckenna1999}. On the negative side however, social media increase the risk of children being confronted with threatening situations including grooming or sexually transgressive behaviour, signals of depression and suicidal thoughts, and cyberbullying. Users are reachable 24/7 and are often able to remain anonymous if desired: this makes social media a convenient way for bullies to target their victims outside the school yard.

With regard to cyberbullying, a number of national and international initiatives have been launched over the past few years to increase children's online safety. Examples include \emph{KiVa}\footnote{\url{http://www.kivaprogram.net/}}, a Finnish cyberbullying prevention programme, the `\emph{Non au harc\`{e}lement}' campaign in France, Belgian governmental initiatives and helplines (e.g.~\emph{clicksafe.be}, \emph{veiligonline.be}, \emph{mediawijs.be}) that provide information about online safety, and so on.

In spite of these efforts, a lot of undesirable and hurtful content remains online.~\cite{Tokunaga2010} analysed a body of quantitative research on cyberbullying and observed cybervictimisation rates among teenagers between 20\% and 40\%.
~\cite{Juvonen2008} focused on 12 to 17 year olds living in the United States and found that no less than 72\% of them had encountered cyberbullying at least once within the year preceding the questionnaire.~\cite{Hinduja2006} surveyed 9 to 26 year olds in the United States, Canada, the United Kingdom and Australia, and found that 29\% of the respondents had ever been victimised online. A study among 2,000 Flemish secondary school students (age 12 to 18) revealed that 11\% of them had been bullied online at least once in the six months preceding the survey \citep{VanCleemput2013}. Finally, the 2014 large-scale EU Kids Online Report~\citep{EUKidsOnline2014} published that 20\% of 11 to 16 year olds had been exposed to hate messages online. In addition, youngsters were 12\% more likely to be exposed to cyberbullying as compared to 2010, clearly demonstrating that cyberbullying is a growing problem.

The prevalence of cybervictimisation depends on the conceptualisation used in describing cyberbullying, but also on research variables such as location and the number and age span of its participants. Nevertheless, the above-mentioned studies demonstrate that online platforms are increasingly used for bullying, which is a cause for concern given its impact. As shown by~\cite{Cowie2013,Fekkes2006,Moore2001}, cyberbullying can have a negative impact on the victim's self-esteem, academic achievement and emotional well-being.~\cite{Price2010} found that self-reported effects of cyberbullying include negative effects on school grades, feelings like sadness, anger, fear, and depression and in extreme cases, cyberbullying could even lead to self-harm and suicidal thoughts. 

The above studies demonstrate that cyberbullying is a serious problem the consequences of which can be dramatic. Successful early detection of cyberbullying attempts is therefore of key importance to youngsters' mental well-being. However, the amount of information on the Web makes it practically unfeasible for moderators to monitor all user-generated content manually. To tackle this problem, intelligent systems are required that process this information in a fast way and automatically signal potential threats. This way, moderators can respond quickly and prevent threatening situations from escalating. According to recent research, teenagers are generally in favour of such automatic monitoring, provided that effective follow-up strategies are formulated, and that privacy and autonomy are guaranteed~\citep{VanRoyen2014}. 

Parental control tools (e.g.~\emph{NetNanny}\footnote{\url{https://www.netnanny.com/}}) already block unsuited or undesirable content and some social networks make use of keyword-based moderation tools (i.e., using lists of profane and insulting words to flag harmful content). However, such approaches typically fail to detect implicit or subtle forms of cyberbullying in which no explicit vocabulary is used. There is therefore a need for intelligent and self-learning systems that can go beyond keyword spotting and hence improve recall of cyberbullying detection.

The ultimate goal of this sort of research is to develop models which could improve manual monitoring for cyberbullying on social networks. We explore the automatic detection of textual signals of cyberbullying, in which it is approached as a complex phenomenon that can be realised in various ways (see Section~\ref{subs:guidelines} for a detailed overview). While a lot of the related research focuses on the detection of cyberbullying `attacks', the present study takes into account a broader range of textual signals of cyberbullying, including posts written by bullies, as well as by victims and bystanders.

We propose a machine learning method to cyberbullying detection by making use of a linear SVM classifier~\citep{Chang2011,Cortes1995} exploiting a varied set of features. To the best of our knowledge, this is the first approach to the annotation of fine-grained text categories related to cyberbullying and the detection of \textbf{signals of cyberbullying events}. It is also the first elaborate research on automatic cyberbullying detection on \textbf{Dutch} social media. For the present experiments, we focus on an English and Dutch ASKfm\footnote{\url{https://ask.fm/}} corpus, but the methodology adopted is language and genre independent, provided there is annotated data available.

The remainder of this paper is structured as follows: the next section presents a theoretic overview and gives an overview of the state of the art in cyberbullying detection, whereas Section~\ref{sect:data} describes the corpus. Next, we present the experimental setup and discuss our experimental results. Finally, Section~\ref{sect:conclusion} concludes this paper and provides perspectives for further research.

\section{Related Research}
\label{sect:rel-research}
Cyberbullying is a widely covered topic in the realm of social sciences and psychology. A fair amount of research has been done on the definition and prevalence of the phenomenon~\citep{Hinduja2012,Livingstone2010,Slonje2008}, the identification of different forms of cyberbullying~\citep{Osullivan2003,Vandebosch2009,Willard2007}, and its consequences~\citep{Cowie2013,Price2010,Smith2008}. In contrast to the efforts made in defining and measuring cyberbullying, the number of studies that focus on its annotation and automatic detection, is limited~\citep{Nadali2013}. Nevertheless, some important advances have been made in the domain over the past few years.

\subsection{A Definition of Cyberbullying}\label{sect:definition}
Many social and psychological studies have worked towards a definition of cyberbullying. A common starting point for conceptualising cyberbullying are definitions of traditional (or \emph{offline}) bullying. Seminal work has been published by \citep{Olweus1993,Nansel2001,Salmivalli1999,Wolak2007}, who describe bullying based on three main criteria, including i) \textbf{intention} (i.e., a bully intends to inflict harm on the victim), ii) \textbf{repetition} (i.e., bullying acts take place repeatedly over time) and iii) a \textbf{power imbalance} between the bully and the victim (i.e., a more powerful bully attacks a less powerful victim). With respect to cyberbullying, a number of definitions are based on the above-mentioned criteria. A popular definition is that of~\citet[p. 376]{Smith2008} which describes cyberbullying as ``an aggressive, intentional act carried out by a group or individual, using electronic forms of contact, repeatedly and over time, against a victim who cannot easily defend him or herself''.

Nevertheless, some studies have underlined the differences between offline and online bullying, and have therefore questioned the relevance of the three criteria to the latter. Besides theoretical objections, a number of practical limitations have been observed. Firstly, while~\cite{Olweus1993} claims intention to be inherent to traditional bullying, this is much harder to ascertain in an online environment. Online conversations lack the signals of a face-to-face interaction like intonation, facial expressions and gestures, which makes them more ambiguous than real-life conversations. The receiver may therefore get the wrong impression that they are being offended or ridiculed~\citep{Vandebosch2009}. Another criterion for bullying that might not hold in online situations, is the power imbalance between bully and victim. Although this can be evident in real life (e.g.~the bully is larger, stronger, older than the victim), it is hard to conceptualise or measure in an online environment. It may be related to technological skills, anonymity or the inability of the victim to get away from the bullying~\cite{Dooley2009,Slonje2008,Vandebosch2008}. Empowering for the bully are also inherent characteristics of the Web: once defamatory or confidential information about a person is made public through the Internet, it is hard, if not impossible, to remove.

Finally, while arguing that repetition is a criterion to distinguish cyberbullying from single acts of aggression,
~\citet{Olweus1993} himself states that such a single aggressive action can be considered bullying under certain circumstances, although it is not entirely clear what these circumstances involve. Accordingly,~\citet{Dooley2009} claim that repetition in cyberbullying is problematic to operationalise, as it is unclear what the consequences are of a single derogatory message on a public page. A single act of aggression or humiliation may result in continued distress and humiliation for the victim if it is shared or liked by multiple perpetrators or read by a large audience.~\citet[p. 26]{Slonje2013} compare this with a `snowball effect': one post may be repeated or distributed by other people so that it becomes out of the control of the initial bully and has larger effects than was originally intended.

Given these arguments, a number of less `strict' definitions of cyberbullying were postulated by among others~\citep{Hinduja2006,Juvonen2008,Tokunaga2010}, where a power imbalance and repetition are not deemed necessary conditions for cyberbullying.

The above paragraphs demonstrate that defining cyberbullying is far from trivial, and varying prevalence rates (cf. Section~\ref{intro}) confirm that a univocal definition of the phenomenon is still lacking in the literature~\citep{Tokunaga2010}. Based on existing conceptualisations, we define cyberbullying as \emph{content that is published online by an individual and that is aggressive or hurtful against a victim}. Based on this definition, an annotation scheme was developed (see~\cite{VanHee2015c}) to signal textual characteristics of cyberbullying, including posts from bullies, as well as reactions by victims and bystanders.

\subsection{Detecting and Preventing Cyberbullying}
As mentioned earlier, although research on cyberbullying detection is more limited than social studies on the phenomenon, some important advances have been made in recent years. In what follows, we present a brief overview of the most important natural language processing approaches to cyberbullying detection.

Although some studies have investigated the effectiveness of rule-based modelling~\citep{Reynolds2011}, the dominant approach to cyberbullying detection involves machine learning. Most machine learning approaches are based on supervised~\citep{Dadvar2014,Dinakar2011,Yin2009} or semi-supervised learning
~\citep{Nahar2014}. The former involves the construction of a classifier based on labeled training data, whereas semi-supervised approaches rely on classifiers that are built from a training corpus containing a small set of labeled and a large set of unlabelled instances (a method that is often used to handle data sparsity). As cyberbullying detection essentially involves the distinction between bullying and non-bullying posts, the problem is generally approached as a binary classification task where the positive class is represented by instances containing (textual) cyberbullying, while the negative class includes instances containing non-cyberbullying or `innocent' text.

A key challenge in cyberbullying research is the availability of suitable data, which is necessary to develop models that characterise cyberbullying. In recent years, only a few datasets have become publicly available for this particular task, such as the training sets provided in the context of the CAW 2.0 workshop\footnote{\url{http://caw2.barcelonamedia.org/}} and more recently, the Twitter Bullying Traces dataset \citep{Sui2015}. As a result, several studies have worked with the former or have constructed their own corpus from social media websites that are prone to bullying content, such as YouTube \citep{Dadvar2014,Dinakar2011}, Formspring\footnote{\url{https://spring.me/}}~\citep{Dinakar2011}, and ASKfm
~\citep{VanHee2015a} (the latter two are social networking sites where users can send each other questions or respond to them). Despite the bottleneck of data availability, existing approaches to cyberbullying detection have shown its potential, and the relevance of automatic text analysis techniques to ensure child safety online has been recognised~\citep{Desmet2014,VanRoyen2016}.

Among the first studies on cyberbullying detection are~\citet{Yin2009,Reynolds2011,Dinakar2011}, who explored the predictive power of $n$-grams (with and without tf-idf weighting), part-of-speech information (e.g.~first and second pronouns), and sentiment information based on profanity lexicons for this task. Similar features were also exploited for the detection of cyberbullying events and fine-grained text categories related to cyberbullying~\citep{VanHee2015a,VanHee2015b}. More recent studies have demonstrated the added value of combining such content-based features with user-based information, such as including users' activities on a social network (i.e., the number of posts), their age, gender, location, number of friends and followers, and so on \citep{Dadvar2014,Nahar2014,Algaradi2016}. Moreover, semantic features have been explored to further improve classification performance of the task. To this end, topic model information~\citep{Xu2012}, as well as semantic relations between $n$-grams (according to a \texttt{Word2Vec} model~\citep{Zhao2016}) have been integrated.

As mentioned earlier, data collection remains a bottleneck in cyberbullying research. Although cyberbullying has been recognised as a serious problem (cf. Section~\ref{intro}), real-world examples are often hard to find in public platforms. Naturally, the vast majority of communications do not contain traces of verbal aggression or transgressive behaviour. When constructing a corpus for machine learning purposes, this results in imbalanced datasets, meaning that one class (e.g.~cyberbullying posts) is much less represented in the corpus than the other (e.g.~non-cyberbullying posts). To tackle this problem, several studies have adopted resampling techniques~\citep{Nahar2014,Algaradi2016,Reynolds2011} that create synthetic minority class examples or reduce the number of negative class examples (i.e., minority class oversampling and majority class undersampling~\citep{Chawla2002}). 

Table~\ref{table:rel-research} presents a number of recent studies on cyberbullying detection, providing insight into the state of the art in cyberbullying research and the contribution of the current research to the domain.

The studies discussed in this section have demonstrated the feasibility of automatic cyberbullying detection in social media data by making use of a varied set of features. Most of them have, however, focussed on cyberbullying `attacks', or posts written by a bully. Moreover, it is not entirely clear if different forms of cyberbullying have been taken into account (e.g.~sexual intimidation or harassment, or psychological threats), in addition to derogatory language or insults.

In the research described in this paper, cyberbullying is considered a complex phenomenon consisting of different forms of harmful behaviour online, which are described in more detail in our annotation scheme~\citep{VanHee2015c}. Purposing to facilitate manual monitoring efforts on social networks, we develop a system that automatically detects signals of cyberbullying, including attacks from bullies, as well as victim and bystander reactions. Similarly,~\citet{Xu2012} investigated bullying traces posted by different author roles (accuser, bully, reporter, victim). However, they collected tweets by using specific keywords (i.e., \emph{bully, bullied} and \emph{bullying}). As a result, their corpus contains many reports or testimonials of a cyberbullying incident (example 1), instead of actual signals that cyberbullying is going on. Moreover, their method implies that cyberbullying-related content devoid of such keywords will not be part of the training corpus.

	\begin{examplenumerate}
		\item `Some tweens got violent on the n train, the one boy got off after blows 2 the chest... Saw him cryin as he walkd away :( bullying not cool' \cite[p. 658]{Xu2012}
	\end{examplenumerate}

For this research, English and Dutch social media data were annotated for different forms of cyberbullying, based on the actors involved in a cyberbullying incident. After preliminary experiments for Dutch~\citep{VanHee2015a,VanHee2015b}, we currently explore the viability of detecting cyberbullying-related posts in Dutch and English social media. To this end, binary classification experiments are performed exploiting a rich feature set and optimised hyperparameters.

\captionsetup{font={footnotesize,sc},justification=centering,labelsep=period}
\begin{table}[]
\scalebox{1}{
\centering
\begin{tabular}{p{3cm} p{1.5cm} p{1.5cm} p{2cm}  p{2.5cm}  p{2cm}}
\hline
& \textbf{Data source} & \textbf{Balanced?} & \textbf{Classifier} & \textbf{Features} & \textbf{Result}\\ \hline \hline
\citep{Reynolds2011} & Formspring & \hspace{5mm} \cmark & J48 & lexical & Recall=62\%, Acc=82\%\\ \hline 
\citep{Dadvar2014} & YouTube & \hspace{5mm} \xmark & SVM (linear) & lexical + semantic + user-based & F$_1$=64\%\\ \hline  
\citep{Dinakar2012} & YouTube & \hspace{5mm} \cmark & SVM (polynomial)& lexical & F$_1$=77\%\\ \hline
\citep{Yin2009} & Kongregate, Slashdot, MySpace & \hspace{5mm} \xmark & SVM (linear) & lexical & F$_1$=48\%\\ \hline 
\citep{Nahar2014} & Kongregate, Slashdot, MySpace & \hspace{5mm} \xmark & semi-supervised fuzzy SVM & lexical + user-based & F$_1$=47\%\\ \hline 
\citep{VanHee2015a} & ASKfm & \hspace{5mm} \xmark & SVM (linear) & lexical & F$_1$=55\% \\ \hline
\citep{Algaradi2016} & Twitter & \hspace{5mm} \xmark & Random Forest + SMOTE & lexical + semantic + user-based + Twitter-based & F$_{1(micro)}$=94\% AUC=94\% \\ \hline
\citep{Xu2012} & Twitter & \hspace{5mm}\cmark & SVM (linear) & lexical & F$_1$=77\% \\ \hline
\citep{Zhao2016} & Twitter & \hspace{5mm}\xmark & SVM (linear) & lexical + semantic & F$_1$=78\% \\
\hline
\end{tabular}}
\caption{State-of-the-art approaches to cyberbullying detection.}
\label{table:rel-research}
\end{table}

\section{Data Collection and Annotation}
\label{sect:data}
To be able to build representative models for cyberbullying, a suitable dataset is required. This section describes the construction of two corpora, English and Dutch, containing social media posts that are manually annotated for cyberbullying according to our fine-grained annotation scheme. This allows us to develop a detection system covering different forms and participants (or \emph{roles}) involved in a cyberbullying event.

\subsection{Data Collection}
Two corpora were constructed by collecting data from the social networking site ASKfm, where users can create profiles and ask or answer questions, with the option of doing so anonymously. ASKfm data typically consists of question-answer pairs published on a user's profile. The data were retrieved by crawling a number of seed profiles using the GNU Wget software\footnote{\url{https://www.gnu.org/software/wget/}} in April and October, 2013. After language filtering (i.e., non-English or non-Dutch content was removed), the experimental corpora comprised 113,698 and 78,387 posts for English and Dutch, respectively.

\subsection{Data Annotation}
Cyberbullying has been a widely covered research topic recently and studies have shed light on direct and indirect types of cyberbullying, implicit and explicit forms, verbal and non-verbal cyberbullying, and so on. This is important from a sociolinguistic point of view, but knowing what cyberbullying involves is also crucial to build models for automatic cyberbullying detection. In the following paragraphs, we present our data annotation guidelines\citep{VanHee2015c} and focus on different types and roles related to the phenomenon. 

\subsection{Types of Cyberbullying}
Cyberbullying research is mainly centered around the conceptualisation, occurrence and prevention of the phenomenon~\citep{Hinduja2012,Livingstone2010,Slonje2008}. Additionally, different forms of cyberbullying have been identified~\citep{Osullivan2003,Price2010,Willard2007} and compared with forms of traditional or offline bullying~\citep{Vandebosch2009}. Like traditional bullying, direct and indirect forms of cyberbullying have been identified. Direct cyberbullying refers to actions in which the victim is directly involved (e.g.~sending a virus-infected file, excluding someone from an online group, insulting and threatening), whereas indirect cyberbullying can take place without awareness of the victim (e.g.~\emph{outing} or publishing confidential information, spreading gossip, creating a hate page on social networking sites)~\citep{Vandebosch2009}.

The present annotation scheme describes some specific textual categories related to cyberbullying, including threats, insults, defensive statements from a victim, encouragements to the harasser, etc. (see Section~\ref{subs:guidelines} for a complete overview). All of these forms were inspired by social studies on cyberbullying\citep{VanCleemput2013,Vandebosch2009} and manual inspection of cyberbullying examples.

\subsection{Roles in Cyberbullying}
Similarly to traditional bullying, cyberbullying involves a number of participants that adopt well-defined roles. Researchers have identified several roles in (cyber)bullying interactions. Although traditional studies on bullying have mainly concentrated on bullies and victims~\citep{Salmivalli1996}, the importance of bystanders in a bullying episode has been acknowledged~\citep{Bastiaensens2014,Salmivalli2010}. Bystanders can support the victim and mitigate the negative effects caused by the bullying~\citep{Salmivalli2010}, especially on social networking sites, where they hold higher intentions to help the victim than in real life conversations~\citep{Bastiaensens2015}. While~\citet{Salmivalli1996} distinguish four different bystanders,~\citet{Vandebosch2006} distinguish three main types: i) bystanders who participate in the bullying, ii) who help or support the victim and iii) those who ignore the bullying. Given that passive bystanders are hard to recognise in online text, only the former two are included in our annotation scheme.

\subsection{Annotation Guidelines}
\label{subs:guidelines}
To operationalise the task of automatic cyberbullying detection, we developed and tested a fine-grained annotation scheme and applied it to our corpora. While a detailed overview of the guidelines is presented in our technical report~\citep{VanHee2015c}, we briefly present the categories and main annotation steps below.

\begin{itemize}
	\item[-]\textbf{Threat/Blackmail:}~expressions containing physical or psychological threats or indications of blackmail.
	\item[-]\textbf{Insult:} expressions meant to hurt or offend the victim.
		\begin{itemize}
			\item[*]\textbf{General insult:} general expressions containing abusive, degrading or offensive language that are meant to insult the addressee.
			\item[*]\textbf{Attacking relatives:} insulting expressions towards relatives or friends of the victim.
			\item[*]\textbf{Discrimination:} expressions of unjust or prejudicial treatment of the victim. Two types of discrimination are distinguished (i.e., sexism and racism). Other forms of discrimination should be categorised as general insults.
		\end{itemize}
	\item[-]\textbf{Curse/Exclusion:} expressions of a wish that some form of adversity or misfortune will befall the victim and expressions that exclude the victim from a conversation or a social group.
	\item[-]\textbf{Defamation:} expressions that reveal confident or defamatory information about the victim to a large public.
	\item[-]\textbf{Sexual Talk:} expressions with a sexual meaning or connotation. A distinction is made between innocent sexual talk and sexual harassment.
	\item[-]\textbf{Defense:} expressions in support of the victim, expressed by the victim himself or by a bystander.
		\begin{itemize}
			\item[*]\textbf{Bystander defense:} expressions by which a bystander shows support for the victim or discourages the harasser from continuing his actions.
			\item[*] \textbf{Victim defense:} assertive or powerless reactions from the victim.
		\end{itemize}
	\item[-]\textbf{Encouragement to the harasser:} expressions in support of the harasser.
	\item[-]\textbf{Other:} expressions that contain any other form of cyberbullying-related behaviour than the ones described here.
\end{itemize}

Based on the literature on role-allocation in cyberbullying episodes~\citep{Salmivalli2011,Vandebosch2006}, four roles are distinguished, including victim, bully, and two types of bystanders.

	\begin{enumerate}
		\item \textbf{Harasser or Bully:} person who initiates the bullying.
		\item \textbf{Victim:} person who is harassed.
		\item \textbf{Bystander-defender:} person who helps the victim and discourages the harasser from continuing his actions.
		\item \textbf{Bystander-assistant:} person who does not initiate, but helps or encourages the harasser.
	\end{enumerate}

Essentially, the annotation scheme describes two levels of annotation. Firstly, the annotators were asked to indicate, at the post level, whether the post under investigation was related to cyberbullying. If the post was considered a signal of cyberbullying, annotators identified the author's role. Secondly, at the subsentence level, the annotators were tasked with the identification of a number of fine-grained text categories related to cyberbullying. More concretely, they identified all text spans corresponding to one of the categories described in the annotation scheme. To provide the annotators with some context, all posts were presented within their original conversation when possible. All annotations were done using the Brat rapid annotation tool~\citep{Stenetorp2012}, some examples of which are presented in Table~\ref{table:ann-ex}.

\captionsetup{font={footnotesize,sc},justification=centering,labelsep=period}
\begin{table*}[htbp]
\centering
 \scalebox{1}{
	\begin{tabular}[t]{p{2.8cm}p{11cm}}
    	\hline
    	\textbf{Annotation} & \textbf{Annotation example} \\
    	\textbf{category} &  \\
    	\hline \hline
    	Threat/blackmail & [I am going to find out who you are \& I swear you are going to regret it.]$^{\textsc{threat}}$ \\ [2ex]
    	Insult & [Kill yourself] $^{\textsc{curs}}$ [you fucking mc slut !!!!]$^{\textsc{gen.~insult}}$ [NO ONE LIKES YOU !!!!!]$^{\textsc{gen.~insult}}$ [You are an ugly useless little whore !!!!] $^{\textsc{gen.~insult}}$ \\ [2ex]
    	Curse/Exclusion & [Fuck you.]$^{\textsc{gen.~insult}}$ [Now shush I don't wanna hear anything.]$^{\textsc{curse~or~exclusion}}$  \\ [2ex]
    	Defamation & [She slept with her ex behind his girlfiends back and she and him had broken up.] $^{\textsc{defamation}}$ \\ [2ex]
    	Sexual Talk & [Naked pic of you now.]$^{\textsc{sexual~harassment}}$  \\ [2ex]
    	Defense & [I would appreciate if you dindn't talk shit about my bestfriend.]$^{\textsc{gen.~victim~defense}}$ He has enough to deal with already.  \\ [2ex]
        Encour. to har. & [She is a massive slut]$^{\textsc{gen.~insult}}$ [i agree with you @user she is!]$^{\textsc{encour.~harasser}}$ [LOL AT HER mate, im on your side]$^{\textsc{encour.~harasser}}$\\ [2ex]
    	\hline
	\end{tabular}}
\caption{Definitions and brat annotation examples of more fine-grained text \\categories related to cyberbullying.}
\label{table:ann-ex}
\end{table*}

\subsection{Annotation Statistics}
\label{subs:ann-stats}
The English and Dutch corpora were independently annotated for cyberbullying by trained linguists. All were Dutch native speakers and English second-language speakers. To demonstrate the validity of our guidelines, inter-annotator agreement scores were calculated using Kappa on a subset of each corpus. Inter-rater agreement for Dutch (2 raters) is calculated using Cohen's Kappa~\citep{Cohen1960}. Fleiss' Kappa~\citep{Fleiss1971} is used for the English corpus ($>$ 2 raters). Kappa scores for the identification of cyberbullying are $\kappa$= 0.69 (Dutch) and $\kappa$= 0.59 (English).

As shown in Table~\ref{table:iaa}, inter-annotator agreement for the identification of the more fine-grained categories for English varies from fair to substantial~\citep{McHugh2012}, except for \emph{defamation}, which appears to be more difficult to recognise. No encouragements to the harasser were present in this subset of the corpus. For Dutch, the inter-annotator agreement is fair to substantial, except for \emph{curse} and \emph{defamation}. Analysis revealed that one of both annotators often annotated the latter as an insult, and in some cases even did not consider it as cyberbullying-related.

\begin{table}[ht]
\centering
 \scalebox{0.94}{
	\begin{tabular}{c c c c c c c c}
	\hline
	 & \textbf{Threat} & \textbf{Insult} & \textbf{Defense} & \textbf{Sexual} & \textbf{Curse/} & \textbf{Defamation} & \textbf{Encouragements} \\
	 & && & \textbf{Talk} & \textbf{Exclusion} & & \textbf{to the harasser} \\ \hline \hline
	 \textbf{English} & 0.65 & 0.63 & 0.45 & 0.38 & 0.58 & 0.15 & N/A \\
	 \textbf{Dutch} & 0.52 & 0.66 & 0.63 & 0.53 & 0.19 & 0.00 & 0.21 \\
	\hline
	\end{tabular}}
\caption{Inter-annotator agreement on the fine-grained categories related to cyberbullying.}
\label{table:iaa}
\end{table}

In short, the inter-rater reliability study shows that the annotation of cyberbullying is not trivial and that more fine-grained categories like \emph{defamation}, \emph{curse} and \emph{encouragements} are sometimes hard to recognise. It appears that defamations were sometimes hard to distinguish from insults, whereas curses and exclusions were sometimes considered insults or threats. The analysis further reveals that encouragements to the harasser are subject to interpretation. Some are straightforward (e.g.~`I agree we should send her hate'), whereas others are subject to the annotator's judgement and interpretation (e.g.~`hahaha', `LOL').

\section{Experimental Setup} \label{expsetup}
\label{sect:exp-setup}
In this paper, we explore the feasibility of automatically recognising signals of cyberbullying. A crucial difference with state-of-the-art approaches to cyberbullying detection is that we aim to model bullying attacks, as well as reactions from victims and bystanders (i.e., all under one binary label `signals of cyberbullying'), since these could likewise indicate that cyberbullying is going on. The experiments described in this paper focus on the detection of such posts, which are signals of a potential cyberbullying \textbf{event} to be further investigated by human moderators.

The English and Dutch corpus contain 113,698 and 78,387 posts, respectively. As shown in Table~\ref{data-stats}, the experimental corpus features a heavily \textbf{imbalanced} class distribution with the large majority of posts not being part of cyberbullying. In classification, this class imbalance can lead to decreased performance. We apply cost-sensitive SVM as a possible hyperparameter in optimisation to counter this. The cost-sensitive SVM reweighs the penalty parameter $C$ of the error term by the inverse class-ratio. This means that misclassifications of the minority positive class are penalised more than classification errors on the majority negative class. Other pre-processing methods to handle data imbalance in classification include feature filtering metrics and data resampling~\citep{he2009learning}. These methods were omitted as they were found to be too computationally expensive given our high-dimensional dataset.

\begin {table} [ht!]
\center
\begin{tabular}{p{2cm} p{2.5cm} p{3cm}}
\hline
 & \textbf{Corpus size} & \textbf{Number {\scriptsize{(ratio)}} of} \\
  &  & \textbf{bullying posts} \\ \hline \hline
 \textbf{English} & 113,698 & 5,375 \scriptsize{(4.73\%)}\\
 \textbf{Dutch} & 78,387 & 5,106 \scriptsize{(6.97\%)}\\
\hline
\end{tabular}
\captionsetup{justification=centering}
\caption{Statistics of the English and Dutch cyberbullying corpus.}
\label{data-stats}
\end{table}

For the automatic detection of cyberbullying, we performed \textbf{binary classification experiments} using a linear kernel support vector machine (SVM) implemented in LIBLINEAR \citep{Fan2008} by making use of Scikit-learn~\citep{Pedregosa2011}, a machine learning library for Python. The motivation behind this is twofold: i) support vector machines (SVMs) have proven to work well for tasks similar to the ones under investigation~\citep{Desmet2014} and ii) LIBLINEAR allows fast training of large-scale data which allow for a linear mapping (which was confirmed after a series of preliminary experiments using LIBSVM with linear, RBF and polynomial kernels).

The classifier was optimised for feature type (cf. Section~\ref{preproc}) and hyperparameter combinations (cf. Table~\ref{hyperparamtable}). Model selection was done using $10$-fold cross validation in grid search over all possible feature types (i.e., groups of similar features, like different orders of $n$-gram bag-of-words features) and hyperparameter configurations.
The best performing hyperparameters are selected by F$_1$-score on the positive class. The winning model is then retrained on all held-in data and subsequently tested on a hold-out test set to assess whether the classifier is over- or under-fitting. The holdout represents a random sample ($10\%$) of all data. The folds were randomly stratified splits over the hold-in class distribution. Testing all feature type combinations is a rudimentary form of feature selection and provides insight into which types of features work best for this particular task.

Feature selection over all individual features was not performed because of the large feature space (NL: 795,072 and EN: 871,296 individual features).~\citet{Hoste2005}, among other researchers, demonstrated the importance of joint optimisation, where feature selection and hyperparameter optimisation are performed simultaneously, since the techniques mutually influence each other.

The optimised models are evaluated against \textbf{two baseline systems}: i) an unoptimised linear-kernel SVM (configured with default parameter settings) based on word $n$-grams only and, ii) a keyword-based system that marks posts as positive for cyberbullying if they contain a word from existing vocabulary lists composed by aggressive language and profanity terms.

\begin{table}
\center
 \scalebox{1}{
 \begin{tabular}{ p{6cm} | p{6cm} }
 \hline
 \textbf{Hyperparameter} & \textbf{Values} \\ [0.5ex]
 \hline\hline
 Penalty of error term $C$ & $1e^{\{-3, -2, ..., 2, 3\}}$ \\ 
 \hline
 Loss function & Hinge, squared hinge\\
 \hline
 Penalty: norm used in penalisation & `l1' (`least absolute deviations') or `l2' (`least squares')\\
 \hline
 Class weight (sets penalty $C$ of class $i$ to weight*$C$) & None or `balanced', i.e., weight inversely proportional to class frequencies\\
 \hline
\end{tabular}}
\captionsetup{justification=centering}
\caption{Hyperparameters in grid-search model selection.}
\label{hyperparamtable}
\end{table}

\subsection{Pre-processing and Feature Engineering}
\label{preproc}
As pre-processing, we applied tokenisation, PoS-tagging and lemmatisation to the data using the LeTs Preprocess Toolkit~\citep{vandeKauter2013}. In supervised learning, a machine learning algorithm takes a set of training instances (of which the label is known) and seeks to build a model that generates a desired prediction for an unseen instance. To enable the model construction, all instances are represented as a vector of features (i.e., inherent characteristics of the data) that contain information that is potentially useful to distinguish cyberbullying from non-cyberbullying content.

We experimentally tested whether cyberbullying events can be recognised automatically by lexical markers in a post. To this end, all posts were represented by a number of information sources (or \emph{features}) including lexical features like bags-of-words, sentiment lexicon features and topic model features, which are described in more detail below. Prior to feature extraction, some data cleaning steps were executed, such as the replacement of hyperlinks and @-replies, removal of superfluous white spaces, and the replacement of abbreviations by their full form (based on an existing mapping dictionary \footnote{\url{http://www.chatslang.com/terms/abbreviations/}}). Additionally, tokenisation was applied before $n$-gram extraction and sentiment lexicon matching, and stemming was applied prior to extracting topic model features.

After pre-processing of the corpus, the following feature types were extracted:

\begin{itemize}
	\item \textbf{Word $n$-gram bag-of-words:} binary features indicating the presence of word unigrams, bigrams and trigrams.
	\item \textbf{Character $n$-gram bag-of-words:} binary features indicating the presence of character bigrams, trigrams and fourgrams (without crossing word boundaries). Character $n$-grams provide some abstraction from the word level and provide robustness to the spelling variation that characterises social media data.
	\item \textbf{Term lists:} one binary feature derived for each one out of six lists, indicating the presence of an item from the list in a post: proper names, `allness' indicators (e.g.~\emph{always}, \emph{everybody}), diminishers (e.g.~\emph{slightly}, \emph{relatively}), intensifiers (e.g.~\emph{absolutely, amazingly}), negation words and aggressive language and profanity words. Person alternation is a binary feature indicating whether the combination of a first and second person pronoun occurs in order to capture interpersonal intent.
	\item \textbf{Subjectivity lexicon features:} positive and negative opinion word ratios, as well as the overall post polarity were calculated using existing sentiment lexicons. For Dutch, we made use of the Duoman~\citep{Jijkoun2009} and Pattern~\citep{DeSmedt2012} lexicons. For English, we included the Hu and Liu opinion lexicon~\citep{Liu2004}, the MPQA lexicon~\citep{Wilson2005}, General Inquirer Sentiment Lexicon~\citep{Stone1966}, AFINN~\citep{Afinn2011}, and MSOL~\citep{Mohammad2009}. For both languages, we included the relative frequency of all 68 psychometric categories in the Linguistic Inquiry and Word Count (LIWC) dictionary for English\citep{Pennebaker2001} and Dutch~\citep{Zijlstra2004}.
	\item \textbf{Topic model features:} by making use of the Gensim topic modelling library~\citep{Rehurek2010}, several LDA~\citep{Blei2003} and LSI~\citep{Deerwester1990} topic models with varying granularity ($k$ = 20, 50, 100 and 200) were trained on data corresponding to each fine-grained category of a cyberbullying event (e.g.~threats, defamations, insults, defenses). The topic models were based on a background corpus (EN: $\pm~1,200,000$ tokens, NL: $\pm~1,400,000$ tokens) scraped with the BootCAT~\citep{Baroni2004} web-corpus toolkit. BootCaT collects ASKfm user profiles using lists of manually determined seed words that are characteristic of the cyberbullying categories.
\end{itemize}

\noindent When applied to the training data, this resulted in \textbf{$871,296$} and \textbf{$795,072$} features for English and Dutch, respectively.

\section{Results}
\label{sect:results}
In this section, we present the results of our experiments on the automatic detection of cyberbullying-related posts in an English (EN) and Dutch (NL) corpus of ASKfm posts. Ten-fold cross-validation was performed in exhaustive grid-search over different feature type and hyperparameter combinations (see Section~\ref{expsetup}). The \emph{unoptimised word $n$-gram-based} classifier and \emph{keyword-matching} system serve as baselines for comparison. Precision, Recall and F$_1$ performance metrics were calculated on the positive class (i.e., `binary averaging'). We also report Area Under the ROC curve (AUC) scores, a performance metric that is more robust to data imbalance than precision, recall and micro-averaged F-score~\citep{fawcett2006introduction}.

\begin {table} [ht!]
\center
\small
\scalebox{0.90}{
\begin{tabular}{l l | l l l l l| l l l l l}
   \hline
   &  \textbf{Feature combination} &\multicolumn{5}{c|}{\textbf{Cross-validation scores}} &\multicolumn{5}{c}{\textbf{Holdout scores}} \\
	& & F$_1$ & P & R & Acc & AUC & F$_1$ & P & R & Acc & AUC \\ \hline \hline
	\textbf{EN} &  &   &   &   &   &   &   &   &   &   &   \\
	& B + C + D + E  & \textbf{64.26}  & 73.32 & 57.19 & 96.97 & 78.07 & 63.69 & 74.13 & 55.82 & 97.21 & 77.47 \\
	& A + B + C & 64.24 & 73.22 & 57.23 & 96.96 & 78.09 & \textbf{64.32} & 74.08 & 56.83 & 97.24 & 77.96 \\
	& A + C + E & 63.84 & 73.21 & 56.59 & 96.94 & 77.78 & 62.94 & 72.82 & 55.42 & 97.14 & 77.24 \\ 
    & word $n$-gram baseline & 58.17 & 67.55 & 51.07 & 96.54 & 74.93 & 59.63 & 69.57 & 52.17 & 96.57 & 75.50 \\ 
    & profanity baseline & 17.17 & 9.61 & 80.14 & 63.73 & 71.53 & 17.61 & 9.90 & 78.51 & 63.79 & 71.34 \\ 
    \hline
	\textbf{NL} &   &  &    &  &   &   &   &   &   &   &  \\
    & A + B + C + E & \textbf{61.20}  & 56.76  & 66.40 & 94.47 & 81.42 & 58.13 & 54.03 & 62.90 & 94.58 & 79.75 \\
	& A + B + C + D + E & 61.03 & 71.55 & 53.20 & 95.53 & 75.86 & \textbf{58.72} & 67.40 & 52.03 & 95.62 & 75.21 \\
	& A + C + E & 60.82 & 71.66 & 52.84 & 95.53 & 75.68 & 58.15 & 67.71 & 50.96 & 95.61 & 74.71 \\
    & word $n$-gram baseline & 50.39 & 67.80 & 40.09 & 94.81 & 69.38 & 49.54 & 64.29 & 40.30 & 95.09 & 69.44 \\ 
    & profanity baseline & 28.46 & 19.24 & 54.66 & 81.99 & 69.28 & 25.13 & 16.73 & 50.53 & 81.99 & 67.26 \\
	\hline
\end{tabular}}
\captionsetup{justification=centering}
\caption{Cross-validated and holdout scores (\%) according to different\\metrics (F$_1$, precision, recall, accuracy and area under the curve)\\for the English and Dutch top 3 combined feature type systems.}
 \label{allresulttable}
\end{table}

\begin {table} [ht!]
\small
\centering
\begin{tabular}{l | l}
	\hline
    A & word $n$-grams \\
    B & subjectivity lexicons \\
    C & character $n$-grams \\
    D & term lists \\
    E & topic models \\
    \hline
\end{tabular}
\captionsetup{justification=centering}
\caption{Feature group mapping (Table~\ref{allresulttable})}
\label{FeatGroupsLegend}
\end{table}

Table~\ref{allresulttable} gives us an indication of which feature type combinations score best and hence contribute most to this task. A total of 31 feature type combinations, each with 28 different hyperparameter sets have been tested. Table~\ref{allresulttable} shows the results for the three best scoring systems by included feature types with optimised hyperparameters. The maximum attained F$_1$-score in cross-validation is 64.26\% for English and 61.20\% for Dutch and shows that the classifier benefits from a variety of feature types. The results on the holdout test set show that the trained systems generalise well on unseen data, indicating little under- or overfitting. The simple keyword-matching baseline system has the lowest performance for both languages even though it obtains high recall for English, suggesting that profane language characterises many cyberbullying-related posts. Feature group and hyperparameter optimisation provides a considerable performance increase over the unoptimised word $n$-gram baseline system. The top-scoring systems for each language do not differ a lot in performance, except the best system for Dutch, which trades recall for precision when compared to the runner-ups.

Table~\ref{onefgresulttable} presents the scores of the (hyperparameter-optimised) single feature type systems, to gain insight into the performance of these feature types when used individually. Analysis of the combined and single feature type sets reveals that \textbf{word $n$-grams, character $n$-grams}, and \textbf{subjectivity lexicons} prove to be strong features for this task. In effect, adding character $n$-grams always improved classification performance for both languages. They likely provide robustness to lexical variation in social media text, as compared to word $n$-grams. While subjectivity lexicons appear to be discriminative features, term lists perform badly on their own as well as in combinations for both languages. This shows once again (cf. profanity baseline) that cyberbullying detection requires more sophisticated information sources than profanity lists. Topic models seem to do badly for both languages on their own, but in combination, they improve Dutch performance consistently. A possible explanation for their varying performance in both languages would be that the topic models trained on the Dutch background corpus are of better quality than the English ones. In effect, a random selection of background corpus texts reveals that the English scrape contains more noisy data (i.e., low word-count posts and non-English posts) than the Dutch data.

\begin {table} [ht!]
\center
\scalebox{0.89}{
\begin{tabular}{l l | l l l l l | l l l l l}
   \hline
    &  \textbf{Feature type} &\multicolumn{5}{c|}{\textbf{Cross-validation scores}}         &\multicolumn{5}{c}{\textbf{holdout scores}} \\
	& & F$_1$ & P & R & Acc &AUC &F$_1$ & P & R & Acc & AUC \\ \hline \hline
    \textbf{EN} &  &   &   &   &   &   &   &   &  &   &  \\
    & word $n$-grams & \textbf{60.09} & 60.49 & 59.69 & 96.22 & 78.87 & \textbf{58.35} & 57.12 & 59.64 & 96.27 & 78.79 \\
    & subjectivity lexicons & 56.82 & 73.32 & 46.38  & 96.64  & 72.77  & 56.16  & 72.61  & 45.78  & 96.87  & 72.50 \\
    & character $n$-grams & 52.69 & 58.70 & 47.80 & 95.91  & 73.06 & 53.33  & 62.37  & 46.59  & 96.43  & 72.65 \\
    & term lists & 40.48 & 38.98 & 42.12  & 94.10 & 69.41 & 39.56 & 39.56  & 39.56  & 94.71  & 68.39 \\
    & topic models & 17.35 & 9.73 & 79.91  & 63.72  & 71.41 & 15.70  & 8.72  & 78.51 & 63.07  & 70.44\\
    \hline
    \textbf{NL} & &  &  &  &  & &  &  &  & &  \\
    & word $n$-grams & \textbf{55.53} & 72.64 & 44.94  & 95.27 & 71.88 & \textbf{54.99} & 70.20 & 45.20  & 95.57 & 71.99 \\
    & subjectivity lexicons & 54.34  & 54.12 & 54.56  & 93.97  & 75.65  & 51.82  & 50.61  & 53.09  & 94.09  & 74.90 \\
    & character $n$-grams & 51.70 & 67.58 & 41.86  & 94.86 & 70.22 & 50.46  & 65.20  & 41.15  & 95.17 &  69.88\\
    & term lists & 28.65  & 19.36  & 55.10 & 81.97  & 69.48 &  25.13 & 16.73 & 50.53 & 81.99 & 67.26  \\
    & topic models & 24.74 & 21.24 & 29.61 & 88.16  & 60.94 & 17.99 & 23.15  & 14.71  & 91.98 & 55.80 \\
    \hline
\end{tabular}}
\captionsetup{justification=centering}
\caption{Cross-validated and holdout scores (\%) according to different\\metrics (F$_1$, precision, recall, accuracy and area under the ROC curve)\\for English and Dutch single feature type systems.}
\label{onefgresulttable}
\end{table}

A shallow \textbf{qualitative analysis} of the classification output provided insight into some of the classification mistakes.

Table~\ref{errortypetable} gives an overview of the error rates per cyberbullying category of the best performing and baseline systems. This could give an indication of which types of bullying the current system has trouble classifying. All categories are always considered positive for cyberbullying (i.e., the error rate equals the false negative rate), except for \emph{Sexual} and \emph{Insult} which can also be negative (in case of harmless sexual talk and `socially acceptable' insulting language like `hi bitches, in for a movie?' the corresponding category was indicated, but the post itself was not annotated as cyberbullying) and \emph{Not cyberbullying}, which is always negative. Error rates often being lowest for the profanity baseline confirms that it performs particularly well in terms of recall (at the expense of precision, see Table~\ref{onefgresulttable}) When looking at the best system for both languages, we see that \emph{Defense} is the hardest category to correctly classify. This should not be a surprise as the category comprises defensive posts from bystanders and victims, which contain less aggressive language than cyberbullying attacks and are often shorter in length than the latter. Assertive defensive posts (i.e., a subcategory of \emph{Defense}) that attack the bully) are, however, more often correctly classified. There are not enough instances of Encouragement for either language in the holdout to be representative. In both languages, threats, curses and incidences of sexual harassment are most easily recognisable, showing (far) lower error rates than the categories \emph{Defamation}, \emph{Defense}, \emph{Encouragements to the harasser}, and \emph{Insult}.

Qualitative error analysis of the English and Dutch predictions reveals that false positives often contain aggressive language directed at a second person, often denoting personal flaws or containing sexual and profanity words. We see that misclassifications are often short posts containing just a few words and that false negatives often lack explicit verbal signs of cyberbullying (e.g.~insulting or profane words) or are ironic (examples 2 and 3). Additionally, we see that cyberbullying posts containing misspellings or grammatical errors and incomplete words are also hard to recognise as such (examples 4 and 5). The Dutch and English data are overall similar with respect to qualitative properties of classification errors. 
     \begin{examplenumerate}
        \item You might want to do some sports ahah x
        \item Look who is there... my thousandth anonymous hater, congratulations!
     	\item ivegot 1 word foryou... yknow whatit is? $\rightarrow$ slut
        \item One word for you: G - A - ...
      \end{examplenumerate}

\begin {table} [ht!]
\center
\small
\scalebox{1}{
\begin{tabular}{l l | l | l l l l l| l l l l l}
   \hline
	& Category & Nr. occurrences & Profanity & Word $n$-gram  & Best system \\
    &  & in holdout & baseline & baseline &  \\ \hline \hline
	\textbf{EN} &  &  &  \\
	& Curse & $n$=109 & 14.68 & 30.28 & 24.77 \\
	& Defamation & $n$=21 & 23.81 & 47.62 & 38.10 \\ 
    & Defense & $n$=165 & 22.42 & 52.12 & 43.64 \\ 
    & Encouragement & $n$=1 & 0.00 & 100.00 & 100.00 \\ 
    & Insult & $n$=345 & 26.67 & 41.74 & 35.94 \\
    & Sexual & $n$=165 & 63.80 & 21.47 & 21,47 \\
    & Threat & $n$=12 & 8.33 & 41.67 & 25.00 \\
    & Not cyberbullying & $n$=10,714 & 36.94 & 1.10 & 0.76 \\
    \hline
	\textbf{NL} &   &   &   \\
    & Curse & $n$=96 & 39.58 & 50.00 & 22.92 \\
	& Defamation & $n$=6 & 100.00 & 66.67 & 33.33 \\ 
    & Defense & $n$=200 & 52.50 & 63.50 & 46.00 \\ 
    & Encouragement & $n$=5 & 40.00 & 60.00 & 40.00 \\ 
    & Insult & $n$=355 & 43.38 & 47.89 & 28.17 \\
    & Sexual & $n$=37 & 37.84 & 21.62 & 27.03 \\
    & Threat & $n$=15 & 33.33 & 46.67 & 20.00 \\
    & Not cyberbullying & $n$=7,295 & 15.63 & 1.23 & 3.07 \\
	\hline
\end{tabular}}
\captionsetup{justification=centering}
\caption{Error rates (\%) per cyberbullying category on holdout for\\English and Dutch systems.}
 \label{errortypetable}
\end{table}
In short, the experiments show that our classifier clearly outperforms both a keyword-based and word $n$-gram baseline. However, analysis of the classifier output reveals that false negatives often lack explicit clues that cyberbullying is going on, indicating that our system might benefit from irony recognition and integrating world knowledge to capture such implicit realisations of cyberbullying.

Given that we present the first elaborate research on detecting signals of cyberbullying regardless of the author role instead of bully posts alone, crude comparison with the state of the art would be irrelevant. We observe, however, that our classifier obtains competitive results compared to~\citet{Dadvar2014,Dinakar2011,Nahar2014,Yin2009,VanHee2015a}.

\section{Conclusions and Future Research}
\label{sect:conclusion}
The goal of the current research was to investigate the automatic detection of cyberbullying-related posts on social media. Given the information overload on the web, manual monitoring for cyberbullying has become unfeasible. Automatic detection of signals of cyberbullying would enhance moderation and allow to respond quickly when necessary.

Cyberbullying research has often focused on detecting cyberbullying `attacks', hence overlooking posts written by victims and bystanders. However, these posts could just as well indicate that cyberbullying is going on. The main contribution of this paper is that it presents a system for detecting \textbf{signals of cyberbullying} on social media, including posts from bullies, victims and bystanders. A manually annotated cyberbullying dataset was created for two languages, which will be made available for public scientific use.
Moreover, while a fair amount of research has been done on cyberbullying detection for English, we believe this is one of the first papers that focus on Dutch as well.

A set of binary classification experiments were conducted to explore the feasibility of automatic cyberbullying detection on social media. In addition, we sought to determine which information sources contribute to this task. Two classifiers were trained on English and Dutch ASKfm data and evaluated on a holdout test of the same genre. Our experiments reveal that the current approach is a promising strategy for detecting signals of cyberbullying in social media data automatically. After feature selection and hyperparameter optimisation, the classifiers achieved an F$_1$-score of 64.32\% and 58.72\% for English and Dutch, respectively. The systems hereby significantly outperformed a keyword and an (unoptimised) $n$-gram baseline. Analysis of the results revealed that false positives often include implicit cyberbullying or offenses through irony, the challenge of which will constitute an important area for future work.

Another interesting direction for future work would be the detection of fine-grained cyberbullying-related categories such as threats, curses and expressions of racism and hate. When applied in a cascaded model, the system could find severe cases of cyberbullying with high precision. This would be particularly interesting for monitoring purposes, since it would allow to prioritise signals of bullying that are in urgent need for manual inspection and follow-up.

Finally, future work will focus on the detection of participants (or \emph{roles}) typically involved in cyberbullying. This would allow to analyse the context of a cyberbullying incident and hence evaluate its severity. When applied as moderation support on online platforms, such a system would allow to provide feedback in function of the recipient (i.e., a bully, victim, or bystander).

\section{Acknowledgment}
The work presented in this paper was carried out in the framework of the AMiCA IWT SBO-project 120007 project, funded by the government Flanders Innovation \& Entrepreneurship (VLAIO) agency.

\bibliographystyle{apacite}
\bibliography{cyberbullyingdetection}

\begin{thebibliography}{}

\bibitem [\protect \citeauthoryear {%
Al-garadi%
, Varathan%
\BCBL {}\ \BBA {} Ravana%
}{%
Al-garadi%
\ \protect \BOthers {.}}{%
{\protect \APACyear {2016}}%
}]{%
Algaradi2016}
\APACinsertmetastar {%
Algaradi2016}%
\begin{APACrefauthors}%
Al-garadi, M\BPBI A.%
, Varathan, K\BPBI D.%
\BCBL {}\ \BBA {} Ravana, S\BPBI D.%
\end{APACrefauthors}%
\unskip\
\newblock
\APACrefYearMonthDay{2016}{}{}.
\newblock
{\BBOQ}\APACrefatitle {{Cybercrime detection in online communications: The
  experimental case of cyberbullying detection in the Twitter network}}
  {{Cybercrime detection in online communications: The experimental case of
  cyberbullying detection in the Twitter network}}.{\BBCQ}
\newblock
\APACjournalVolNumPages{{Computers in Human Behavior}}{63}{}{433--443}.
\newblock
\begin{APACrefDOI} \doi{http://dx.doi.org/10.1016/j.chb.2016.05.051}
  \end{APACrefDOI}
\PrintBackRefs{\CurrentBib}

\bibitem [\protect \citeauthoryear {%
Baroni%
\ \BBA {} Bernardini%
}{%
Baroni%
\ \BBA {} Bernardini%
}{%
{\protect \APACyear {2004}}%
}]{%
Baroni2004}
\APACinsertmetastar {%
Baroni2004}%
\begin{APACrefauthors}%
Baroni, M.%
\BCBT {}\ \BBA {} Bernardini, S.%
\end{APACrefauthors}%
\unskip\
\newblock
\APACrefYearMonthDay{2004}{}{}.
\newblock
{\BBOQ}\APACrefatitle {{BootCaT: Bootstrapping Corpora and Terms from the Web}}
  {{BootCaT: Bootstrapping Corpora and Terms from the Web}}.{\BBCQ}
\newblock
\BIn{} \APACrefbtitle {{Proceedings of the Fourth International Conference on
  Language Resources and Evaluation}} {{Proceedings of the Fourth International
  Conference on Language Resources and Evaluation}}\ (\BPGS\ 1313--1316).
\PrintBackRefs{\CurrentBib}

\bibitem [\protect \citeauthoryear {%
Bastiaensens%
\ \protect \BOthers {.}}{%
Bastiaensens%
\ \protect \BOthers {.}}{%
{\protect \APACyear {2014}}%
}]{%
Bastiaensens2014}
\APACinsertmetastar {%
Bastiaensens2014}%
\begin{APACrefauthors}%
Bastiaensens, S.%
, Vandebosch, H.%
, Poels, K.%
, Van~Cleemput, K.%
, DeSmet, A.%
\BCBL {}\ \BBA {} De~Bourdeaudhuij, I.%
\end{APACrefauthors}%
\unskip\
\newblock
\APACrefYearMonthDay{2014}{}{}.
\newblock
{\BBOQ}\APACrefatitle {{Cyberbullying on social network sites. An experimental
  study into bystanders' behavioural intentions to help the victim or reinforce
  the bully}} {{Cyberbullying on social network sites. An experimental study
  into bystanders' behavioural intentions to help the victim or reinforce the
  bully}}.{\BBCQ}
\newblock
\APACjournalVolNumPages{{Computers in Human Behavior}}{31}{}{259--271}.
\PrintBackRefs{\CurrentBib}

\bibitem [\protect \citeauthoryear {%
Bastiaensens%
\ \protect \BOthers {.}}{%
Bastiaensens%
\ \protect \BOthers {.}}{%
{\protect \APACyear {2015}}%
}]{%
Bastiaensens2015}
\APACinsertmetastar {%
Bastiaensens2015}%
\begin{APACrefauthors}%
Bastiaensens, S.%
, Vandebosch, H.%
, Poels, K.%
, Van~Cleemput, K.%
, DeSmet, A.%
\BCBL {}\ \BBA {} De~Bourdeaudhuij, I.%
\end{APACrefauthors}%
\unskip\
\newblock
\APACrefYearMonthDay{2015}{}{}.
\newblock
{\BBOQ}\APACrefatitle {{`Can I afford to help?' How affordances of
  communication modalities guide bystanders' helping intentions towards
  harassment on social network sites}} {{`Can I afford to help?' How
  affordances of communication modalities guide bystanders' helping intentions
  towards harassment on social network sites}}.{\BBCQ}
\newblock
\APACjournalVolNumPages{{Behaviour \& Information
  Technology}}{34}{4}{425--435}.
\newblock
\begin{APACrefDOI} \doi{10.1080/0144929X.2014.983979} \end{APACrefDOI}
\PrintBackRefs{\CurrentBib}

\bibitem [\protect \citeauthoryear {%
Blei%
, Ng%
\BCBL {}\ \BBA {} Jordan%
}{%
Blei%
\ \protect \BOthers {.}}{%
{\protect \APACyear {2003}}%
}]{%
Blei2003}
\APACinsertmetastar {%
Blei2003}%
\begin{APACrefauthors}%
Blei, D\BPBI M.%
, Ng, A\BPBI Y.%
\BCBL {}\ \BBA {} Jordan, M\BPBI I.%
\end{APACrefauthors}%
\unskip\
\newblock
\APACrefYearMonthDay{2003}{}{}.
\newblock
{\BBOQ}\APACrefatitle {{Latent dirichlet allocation}} {{Latent dirichlet
  allocation}}.{\BBCQ}
\newblock
\APACjournalVolNumPages{{Journal of Machine Learning
  Research}}{3}{}{993--1022}.
\PrintBackRefs{\CurrentBib}

\bibitem [\protect \citeauthoryear {%
Chang%
\ \BBA {} Lin%
}{%
Chang%
\ \BBA {} Lin%
}{%
{\protect \APACyear {2011}}%
}]{%
Chang2011}
\APACinsertmetastar {%
Chang2011}%
\begin{APACrefauthors}%
Chang, C\BHBI C.%
\BCBT {}\ \BBA {} Lin, C\BHBI J.%
\end{APACrefauthors}%
\unskip\
\newblock
\APACrefYearMonthDay{2011}{}{}.
\newblock
{\BBOQ}\APACrefatitle {{LIBSVM: A Library for Support Vector Machines}}
  {{LIBSVM: A Library for Support Vector Machines}}.{\BBCQ}
\newblock
\APACjournalVolNumPages{{ACM Transactions on Intelligent Systems and Technology
  (TIST)}}{2}{3}{27:1--27:27}.
\newblock
\begin{APACrefDOI} \doi{10.1145/1961189.1961199} \end{APACrefDOI}
\PrintBackRefs{\CurrentBib}

\bibitem [\protect \citeauthoryear {%
Chawla%
, Bowyer%
, Hall%
\BCBL {}\ \BBA {} Kegelmeyer%
}{%
Chawla%
\ \protect \BOthers {.}}{%
{\protect \APACyear {2002}}%
}]{%
Chawla2002}
\APACinsertmetastar {%
Chawla2002}%
\begin{APACrefauthors}%
Chawla, N\BPBI V.%
, Bowyer, K\BPBI W.%
, Hall, L\BPBI O.%
\BCBL {}\ \BBA {} Kegelmeyer, P\BPBI W.%
\end{APACrefauthors}%
\unskip\
\newblock
\APACrefYearMonthDay{2002}{}{}.
\newblock
{\BBOQ}\APACrefatitle {{SMOTE: Synthetic Minority Over-sampling TEchnique}}
  {{SMOTE: Synthetic Minority Over-sampling TEchnique}}.{\BBCQ}
\newblock
\APACjournalVolNumPages{Journal of Artificial Intelligence Research
  (JAIR)}{16}{}{321--357}.
\PrintBackRefs{\CurrentBib}

\bibitem [\protect \citeauthoryear {%
Cohen%
}{%
Cohen%
}{%
{\protect \APACyear {1960}}%
}]{%
Cohen1960}
\APACinsertmetastar {%
Cohen1960}%
\begin{APACrefauthors}%
Cohen, J.%
\end{APACrefauthors}%
\unskip\
\newblock
\APACrefYearMonthDay{1960}{}{}.
\newblock
{\BBOQ}\APACrefatitle {{A coefficient of agreement for nominal scales}} {{A
  coefficient of agreement for nominal scales}}.{\BBCQ}
\newblock
\APACjournalVolNumPages{{Educational and Psychological
  Measurement}}{20}{1}{37--46}.
\PrintBackRefs{\CurrentBib}

\bibitem [\protect \citeauthoryear {%
Cortes%
\ \BBA {} Vapnik%
}{%
Cortes%
\ \BBA {} Vapnik%
}{%
{\protect \APACyear {1995}}%
}]{%
Cortes1995}
\APACinsertmetastar {%
Cortes1995}%
\begin{APACrefauthors}%
Cortes, C.%
\BCBT {}\ \BBA {} Vapnik, V.%
\end{APACrefauthors}%
\unskip\
\newblock
\APACrefYearMonthDay{1995}{}{}.
\newblock
{\BBOQ}\APACrefatitle {{Support-Vector Networks}} {{Support-Vector
  Networks}}.{\BBCQ}
\newblock
\APACjournalVolNumPages{{Machine Learning}}{20}{3}{273--297}.
\newblock
\begin{APACrefDOI} \doi{10.1023/A:1022627411411} \end{APACrefDOI}
\PrintBackRefs{\CurrentBib}

\bibitem [\protect \citeauthoryear {%
Cowie%
}{%
Cowie%
}{%
{\protect \APACyear {2013}}%
}]{%
Cowie2013}
\APACinsertmetastar {%
Cowie2013}%
\begin{APACrefauthors}%
Cowie, H.%
\end{APACrefauthors}%
\unskip\
\newblock
\APACrefYearMonthDay{2013}{}{}.
\newblock
{\BBOQ}\APACrefatitle {{Cyberbullying and its impact on young people's
  emotional health and well-being}} {{Cyberbullying and its impact on young
  people's emotional health and well-being}}.{\BBCQ}
\newblock
\APACjournalVolNumPages{{The Psychiatrist}}{37}{5}{167--170}.
\newblock
\begin{APACrefDOI} \doi{10.1192/pb.bp.112.040840} \end{APACrefDOI}
\PrintBackRefs{\CurrentBib}

\bibitem [\protect \citeauthoryear {%
Dadvar%
}{%
Dadvar%
}{%
{\protect \APACyear {2014}}%
}]{%
Dadvar2014}
\APACinsertmetastar {%
Dadvar2014}%
\begin{APACrefauthors}%
Dadvar, M.%
\end{APACrefauthors}%
\unskip\
\newblock
\APACrefYear{2014}.
\unskip\
\newblock
\APACrefbtitle {{Experts and machines united against cyberbullying}} {{Experts
  and machines united against cyberbullying}}\ \APACtypeAddressSchool {PhD
  thesis}{}{}.
\unskip\
\newblock
\APACaddressSchool {}{University of Twente}.
\PrintBackRefs{\CurrentBib}

\bibitem [\protect \citeauthoryear {%
Deerwester%
, Dumais%
, Furnas%
, Landauer%
\BCBL {}\ \BBA {} Harshman%
}{%
Deerwester%
\ \protect \BOthers {.}}{%
{\protect \APACyear {1990}}%
}]{%
Deerwester1990}
\APACinsertmetastar {%
Deerwester1990}%
\begin{APACrefauthors}%
Deerwester, S.%
, Dumais, S\BPBI T.%
, Furnas, G\BPBI W.%
, Landauer, T\BPBI K.%
\BCBL {}\ \BBA {} Harshman, R.%
\end{APACrefauthors}%
\unskip\
\newblock
\APACrefYearMonthDay{1990}{}{}.
\newblock
{\BBOQ}\APACrefatitle {{Indexing by latent semantic analysis}} {{Indexing by
  latent semantic analysis}}.{\BBCQ}
\newblock
\APACjournalVolNumPages{{Journal of the American Society for Information
  Science}}{41}{}{391--407}.
\PrintBackRefs{\CurrentBib}

\bibitem [\protect \citeauthoryear {%
De~Smedt%
\ \BBA {} Daelemans%
}{%
De~Smedt%
\ \BBA {} Daelemans%
}{%
{\protect \APACyear {2012}}%
}]{%
DeSmedt2012}
\APACinsertmetastar {%
DeSmedt2012}%
\begin{APACrefauthors}%
De~Smedt, T.%
\BCBT {}\ \BBA {} Daelemans, W.%
\end{APACrefauthors}%
\unskip\
\newblock
\APACrefYearMonthDay{2012}{}{}.
\newblock
{\BBOQ}\APACrefatitle {{``Vreselijk mooi!" (``Terribly Beautiful!"): A
  Subjectivity Lexicon for Dutch Adjectives}} {{``Vreselijk mooi!" (``Terribly
  Beautiful!"): A Subjectivity Lexicon for Dutch Adjectives}}.{\BBCQ}
\newblock
\BIn{} \APACrefbtitle {{Proceedings of the Eight International Conference on
  Language Resources and Evaluation}} {{Proceedings of the Eight International
  Conference on Language Resources and Evaluation}}\ (\BPG~3568-3572).
\newblock
\APACaddressPublisher{Istanbul, Turkey}{}.
\PrintBackRefs{\CurrentBib}

\bibitem [\protect \citeauthoryear {%
Desmet%
}{%
Desmet%
}{%
{\protect \APACyear {2014}}%
}]{%
Desmet2014}
\APACinsertmetastar {%
Desmet2014}%
\begin{APACrefauthors}%
Desmet, B.%
\end{APACrefauthors}%
\unskip\
\newblock
\APACrefYear{2014}.
\unskip\
\newblock
\APACrefbtitle {{Finding the online cry for help: automatic text classification
  for suicide prevention}} {{Finding the online cry for help: automatic text
  classification for suicide prevention}}\ \APACtypeAddressSchool {PhD
  thesis}{}{}.
\unskip\
\newblock
\APACaddressSchool {}{Ghent University}.
\PrintBackRefs{\CurrentBib}

\bibitem [\protect \citeauthoryear {%
Dinakar%
, Jones%
, Havasi%
, Lieberman%
\BCBL {}\ \BBA {} Picard%
}{%
Dinakar%
\ \protect \BOthers {.}}{%
{\protect \APACyear {2012}}%
}]{%
Dinakar2012}
\APACinsertmetastar {%
Dinakar2012}%
\begin{APACrefauthors}%
Dinakar, K.%
, Jones, B.%
, Havasi, C.%
, Lieberman, H.%
\BCBL {}\ \BBA {} Picard, R.%
\end{APACrefauthors}%
\unskip\
\newblock
\APACrefYearMonthDay{2012}{}{}.
\newblock
{\BBOQ}\APACrefatitle {{Common Sense Reasoning for Detection, Prevention, and
  Mitigation of Cyberbullying}} {{Common Sense Reasoning for Detection,
  Prevention, and Mitigation of Cyberbullying}}.{\BBCQ}
\newblock
\APACjournalVolNumPages{{ACM Transactions on Interactive Intelligent
  Systems}}{2}{3}{18:1--18:30}.
\PrintBackRefs{\CurrentBib}

\bibitem [\protect \citeauthoryear {%
Dinakar%
, Reichart%
\BCBL {}\ \BBA {} Lieberman%
}{%
Dinakar%
\ \protect \BOthers {.}}{%
{\protect \APACyear {2011}}%
}]{%
Dinakar2011}
\APACinsertmetastar {%
Dinakar2011}%
\begin{APACrefauthors}%
Dinakar, K.%
, Reichart, R.%
\BCBL {}\ \BBA {} Lieberman, H.%
\end{APACrefauthors}%
\unskip\
\newblock
\APACrefYearMonthDay{2011}{}{}.
\newblock
{\BBOQ}\APACrefatitle {{Modeling the Detection of Textual Cyberbullying}}
  {{Modeling the Detection of Textual Cyberbullying}}.{\BBCQ}
\newblock
\BIn{} \APACrefbtitle {{The Social Mobile Web}} {{The Social Mobile Web}}\
  (\BVOL\ WS-11-02, \BPGS\ 11--17).
\newblock
\APACaddressPublisher{}{AAAI}.
\PrintBackRefs{\CurrentBib}

\bibitem [\protect \citeauthoryear {%
Dooley%
\ \BBA {} Cross%
}{%
Dooley%
\ \BBA {} Cross%
}{%
{\protect \APACyear {2009}}%
}]{%
Dooley2009}
\APACinsertmetastar {%
Dooley2009}%
\begin{APACrefauthors}%
Dooley, J\BPBI J.%
\BCBT {}\ \BBA {} Cross, D.%
\end{APACrefauthors}%
\unskip\
\newblock
\APACrefYearMonthDay{2009}{}{}.
\newblock
{\BBOQ}\APACrefatitle {{Cyberbullying versus face-to-face bullying: A review of
  the similarities and differences}} {{Cyberbullying versus face-to-face
  bullying: A review of the similarities and differences}}.{\BBCQ}
\newblock
\APACjournalVolNumPages{{Journal of Psychology}}{217}{}{182--188}.
\PrintBackRefs{\CurrentBib}

\bibitem [\protect \citeauthoryear {%
Fan%
, Chang%
, Hsieh%
, Wang%
\BCBL {}\ \BBA {} Lin%
}{%
Fan%
\ \protect \BOthers {.}}{%
{\protect \APACyear {2008}}%
}]{%
Fan2008}
\APACinsertmetastar {%
Fan2008}%
\begin{APACrefauthors}%
Fan, R\BHBI E.%
, Chang, K\BHBI W.%
, Hsieh, C\BHBI J.%
, Wang, X\BHBI R.%
\BCBL {}\ \BBA {} Lin, C\BHBI J.%
\end{APACrefauthors}%
\unskip\
\newblock
\APACrefYearMonthDay{2008}{}{}.
\newblock
{\BBOQ}\APACrefatitle {{LIBLINEAR: A Library for Large Linear Classification}}
  {{LIBLINEAR: A Library for Large Linear Classification}}.{\BBCQ}
\newblock
\APACjournalVolNumPages{{Journal of Machine Learning
  Research}}{9}{}{1871--1874}.
\PrintBackRefs{\CurrentBib}

\bibitem [\protect \citeauthoryear {%
Fawcett%
}{%
Fawcett%
}{%
{\protect \APACyear {2006}}%
}]{%
fawcett2006introduction}
\APACinsertmetastar {%
fawcett2006introduction}%
\begin{APACrefauthors}%
Fawcett, T.%
\end{APACrefauthors}%
\unskip\
\newblock
\APACrefYearMonthDay{2006}{}{}.
\newblock
{\BBOQ}\APACrefatitle {An introduction to ROC analysis} {An introduction to roc
  analysis}.{\BBCQ}
\newblock
\APACjournalVolNumPages{Pattern recognition letters}{27}{8}{861--874}.
\PrintBackRefs{\CurrentBib}

\bibitem [\protect \citeauthoryear {%
Fekkes%
, Pijpers%
, Fredriks%
, Vogels%
\BCBL {}\ \BBA {} Verloove-Vanhorick%
}{%
Fekkes%
\ \protect \BOthers {.}}{%
{\protect \APACyear {2006}}%
}]{%
Fekkes2006}
\APACinsertmetastar {%
Fekkes2006}%
\begin{APACrefauthors}%
Fekkes, M.%
, Pijpers, F\BPBI I.%
, Fredriks, A\BPBI M.%
, Vogels, T.%
\BCBL {}\ \BBA {} Verloove-Vanhorick, S\BPBI P.%
\end{APACrefauthors}%
\unskip\
\newblock
\APACrefYearMonthDay{2006}{}{}.
\newblock
{\BBOQ}\APACrefatitle {{Do Bullied Children Get Ill, or Do Ill Children Get
  Bullied? A Prospective Cohort Study on the Relationship Between Bullying and
  Health-Related Symptoms}} {{Do Bullied Children Get Ill, or Do Ill Children
  Get Bullied? A Prospective Cohort Study on the Relationship Between Bullying
  and Health-Related Symptoms}}.{\BBCQ}
\newblock
\APACjournalVolNumPages{{Pediatrics}}{117}{5}{1568--1574}.
\newblock
\begin{APACrefDOI} \doi{10.1542/peds.2005-0187} \end{APACrefDOI}
\PrintBackRefs{\CurrentBib}

\bibitem [\protect \citeauthoryear {%
Fleiss%
}{%
Fleiss%
}{%
{\protect \APACyear {1971}}%
}]{%
Fleiss1971}
\APACinsertmetastar {%
Fleiss1971}%
\begin{APACrefauthors}%
Fleiss, J\BPBI L.%
\end{APACrefauthors}%
\unskip\
\newblock
\APACrefYearMonthDay{1971}{}{}.
\newblock
{\BBOQ}\APACrefatitle {{Measuring nominal scale agreement among many raters}}
  {{Measuring nominal scale agreement among many raters}}.{\BBCQ}
\newblock
\APACjournalVolNumPages{{Psychological Bulletin}}{76}{5}{378--382}.
\PrintBackRefs{\CurrentBib}

\bibitem [\protect \citeauthoryear {%
Gross%
, Juvonen%
\BCBL {}\ \BBA {} Gable%
}{%
Gross%
\ \protect \BOthers {.}}{%
{\protect \APACyear {2002}}%
}]{%
Gross2002}
\APACinsertmetastar {%
Gross2002}%
\begin{APACrefauthors}%
Gross, E\BPBI F.%
, Juvonen, J.%
\BCBL {}\ \BBA {} Gable, S\BPBI L.%
\end{APACrefauthors}%
\unskip\
\newblock
\APACrefYearMonthDay{2002}{}{}.
\newblock
{\BBOQ}\APACrefatitle {{Internet Use and Well-Being in Adolescence}} {{Internet
  Use and Well-Being in Adolescence}}.{\BBCQ}
\newblock
\APACjournalVolNumPages{{Journal of Social Issues}}{58}{1}{75--90}.
\PrintBackRefs{\CurrentBib}

\bibitem [\protect \citeauthoryear {%
He%
\ \BBA {} Garcia%
}{%
He%
\ \BBA {} Garcia%
}{%
{\protect \APACyear {2009}}%
}]{%
he2009learning}
\APACinsertmetastar {%
he2009learning}%
\begin{APACrefauthors}%
He, H.%
\BCBT {}\ \BBA {} Garcia, E\BPBI A.%
\end{APACrefauthors}%
\unskip\
\newblock
\APACrefYearMonthDay{2009}{}{}.
\newblock
{\BBOQ}\APACrefatitle {Learning from imbalanced data} {Learning from imbalanced
  data}.{\BBCQ}
\newblock
\APACjournalVolNumPages{IEEE Transactions on knowledge and data
  engineering}{21}{9}{1263--1284}.
\PrintBackRefs{\CurrentBib}

\bibitem [\protect \citeauthoryear {%
Hinduja%
\ \BBA {} Patchin%
}{%
Hinduja%
\ \BBA {} Patchin%
}{%
{\protect \APACyear {2006}}%
}]{%
Hinduja2006}
\APACinsertmetastar {%
Hinduja2006}%
\begin{APACrefauthors}%
Hinduja, S.%
\BCBT {}\ \BBA {} Patchin, J\BPBI W.%
\end{APACrefauthors}%
\unskip\
\newblock
\APACrefYearMonthDay{2006}{}{}.
\newblock
{\BBOQ}\APACrefatitle {{Bullies Move Beyond the Schoolyard: A Preliminary Look
  at Cyberbullying}} {{Bullies Move Beyond the Schoolyard: A Preliminary Look
  at Cyberbullying}}.{\BBCQ}
\newblock
\APACjournalVolNumPages{{Youth Violence And Juvenile Justice}}{4}{2}{148--169}.
\PrintBackRefs{\CurrentBib}

\bibitem [\protect \citeauthoryear {%
Hinduja%
\ \BBA {} Patchin%
}{%
Hinduja%
\ \BBA {} Patchin%
}{%
{\protect \APACyear {2012}}%
}]{%
Hinduja2012}
\APACinsertmetastar {%
Hinduja2012}%
\begin{APACrefauthors}%
Hinduja, S.%
\BCBT {}\ \BBA {} Patchin, J\BPBI W.%
\end{APACrefauthors}%
\unskip\
\newblock
\APACrefYearMonthDay{2012}{}{}.
\newblock
{\BBOQ}\APACrefatitle {{Cyberbullying: Neither an epidemic nor a rarity}}
  {{Cyberbullying: Neither an epidemic nor a rarity}}.{\BBCQ}
\newblock
\APACjournalVolNumPages{{European Journal of Developmental
  Psychology}}{9}{5}{539--543}.
\PrintBackRefs{\CurrentBib}

\bibitem [\protect \citeauthoryear {%
Hoste%
}{%
Hoste%
}{%
{\protect \APACyear {2005}}%
}]{%
Hoste2005}
\APACinsertmetastar {%
Hoste2005}%
\begin{APACrefauthors}%
Hoste, V.%
\end{APACrefauthors}%
\unskip\
\newblock
\APACrefYear{2005}.
\unskip\
\newblock
\APACrefbtitle {{Optimization Issues in Machine Learning of Coreference
  Resolution}} {{Optimization Issues in Machine Learning of Coreference
  Resolution}}\ \APACtypeAddressSchool {PhD thesis}{}{}.
\unskip\
\newblock
\APACaddressSchool {}{Antwerp University}.
\PrintBackRefs{\CurrentBib}

\bibitem [\protect \citeauthoryear {%
Hu%
\ \BBA {} Liu%
}{%
Hu%
\ \BBA {} Liu%
}{%
{\protect \APACyear {2004}}%
}]{%
Liu2004}
\APACinsertmetastar {%
Liu2004}%
\begin{APACrefauthors}%
Hu, M.%
\BCBT {}\ \BBA {} Liu, B.%
\end{APACrefauthors}%
\unskip\
\newblock
\APACrefYearMonthDay{2004}{}{}.
\newblock
{\BBOQ}\APACrefatitle {{Mining and summarizing customer reviews}} {{Mining and
  summarizing customer reviews}}.{\BBCQ}
\newblock
\BIn{} \APACrefbtitle {{Proceedings of the 10th ACM SIGKDD international
  conference on Knowledge discovery and data mining}} {{Proceedings of the 10th
  ACM SIGKDD international conference on Knowledge discovery and data mining}}\
  (\BPGS\ 168--177).
\newblock
\APACaddressPublisher{}{ACM}.
\PrintBackRefs{\CurrentBib}

\bibitem [\protect \citeauthoryear {%
Jijkoun%
\ \BBA {} Hofmann%
}{%
Jijkoun%
\ \BBA {} Hofmann%
}{%
{\protect \APACyear {2009}}%
}]{%
Jijkoun2009}
\APACinsertmetastar {%
Jijkoun2009}%
\begin{APACrefauthors}%
Jijkoun, V.%
\BCBT {}\ \BBA {} Hofmann, K.%
\end{APACrefauthors}%
\unskip\
\newblock
\APACrefYearMonthDay{2009}{}{}.
\newblock
{\BBOQ}\APACrefatitle {{Generating a Non-English Subjectivity Lexicon:
  Relations That Matter}} {{Generating a Non-English Subjectivity Lexicon:
  Relations That Matter}}.{\BBCQ}
\newblock
\BIn{} \APACrefbtitle {{Proceedings of the 12th Conference of the European
  Chapter of the Association for Computational Linguistics}} {{Proceedings of
  the 12th Conference of the European Chapter of the Association for
  Computational Linguistics}}\ (\BPG~398-405).
\newblock
\APACaddressPublisher{Stroudsburg, PA, USA}{}.
\PrintBackRefs{\CurrentBib}

\bibitem [\protect \citeauthoryear {%
Juvonen%
\ \BBA {} Gross%
}{%
Juvonen%
\ \BBA {} Gross%
}{%
{\protect \APACyear {2008}}%
}]{%
Juvonen2008}
\APACinsertmetastar {%
Juvonen2008}%
\begin{APACrefauthors}%
Juvonen, J.%
\BCBT {}\ \BBA {} Gross, E\BPBI F.%
\end{APACrefauthors}%
\unskip\
\newblock
\APACrefYearMonthDay{2008}{}{}.
\newblock
{\BBOQ}\APACrefatitle {{Extending the school grounds? -- Bullying experiences
  in cyberspace}} {{Extending the school grounds? -- Bullying experiences in
  cyberspace}}.{\BBCQ}
\newblock
\APACjournalVolNumPages{{Journal of School Health}}{78}{9}{496--505}.
\PrintBackRefs{\CurrentBib}

\bibitem [\protect \citeauthoryear {%
Livingstone%
, Haddon%
, G\"{o}rzig%
\BCBL {}\ \BBA {} \'{O}lafsson%
}{%
Livingstone%
\ \protect \BOthers {.}}{%
{\protect \APACyear {2010}}%
}]{%
Livingstone2010}
\APACinsertmetastar {%
Livingstone2010}%
\begin{APACrefauthors}%
Livingstone, S.%
, Haddon, L.%
, G\"{o}rzig, A.%
\BCBL {}\ \BBA {} \'{O}lafsson, K.%
\end{APACrefauthors}%
\unskip\
\newblock
\APACrefYearMonthDay{2010}{}{}.
\newblock
\APACrefbtitle {{Risks and safety on the internet: The perspective of European
  children. Initial Findings}.} {{Risks and safety on the internet: The
  perspective of European children. Initial Findings}.}
\newblock
\APACaddressPublisher{}{London: EU Kids Online}.
\PrintBackRefs{\CurrentBib}

\bibitem [\protect \citeauthoryear {%
Livingstone%
, Kirwil%
, Ponte%
\BCBL {}\ \BBA {} Staksrud%
}{%
Livingstone%
\ \protect \BOthers {.}}{%
{\protect \APACyear {2013}}%
}]{%
Livingstone2013}
\APACinsertmetastar {%
Livingstone2013}%
\begin{APACrefauthors}%
Livingstone, S.%
, Kirwil, L.%
, Ponte, C.%
\BCBL {}\ \BBA {} Staksrud, E.%
\end{APACrefauthors}%
\unskip\
\newblock
\APACrefYearMonthDay{2013}{}{}.
\newblock
\APACrefbtitle {{In their own words: what bothers children online?}} {{In their
  own words: what bothers children online?}}
\newblock
\APACaddressPublisher{}{London: EU Kids Online}.
\PrintBackRefs{\CurrentBib}

\bibitem [\protect \citeauthoryear {%
McHugh%
}{%
McHugh%
}{%
{\protect \APACyear {2012}}%
}]{%
McHugh2012}
\APACinsertmetastar {%
McHugh2012}%
\begin{APACrefauthors}%
McHugh, M\BPBI L.%
\end{APACrefauthors}%
\unskip\
\newblock
\APACrefYearMonthDay{2012}{}{}.
\newblock
{\BBOQ}\APACrefatitle {{Interrater reliability: the kappa statistic}}
  {{Interrater reliability: the kappa statistic}}.{\BBCQ}
\newblock
\APACjournalVolNumPages{{Biochemia Medica}}{22}{3}{276--282}.
\PrintBackRefs{\CurrentBib}

\bibitem [\protect \citeauthoryear {%
Mckenna%
\ \BBA {} Bargh%
}{%
Mckenna%
\ \BBA {} Bargh%
}{%
{\protect \APACyear {1999}}%
}]{%
Mckenna1999}
\APACinsertmetastar {%
Mckenna1999}%
\begin{APACrefauthors}%
Mckenna, K\BPBI Y.%
\BCBT {}\ \BBA {} Bargh, J\BPBI A.%
\end{APACrefauthors}%
\unskip\
\newblock
\APACrefYearMonthDay{1999}{}{}.
\newblock
{\BBOQ}\APACrefatitle {{Plan 9 From Cyberspace: The Implications of the
  Internet for Personality and Social Psychology.}} {{Plan 9 From Cyberspace:
  The Implications of the Internet for Personality and Social
  Psychology.}}{\BBCQ}
\newblock
\APACjournalVolNumPages{Personality \& Social Psychology Review}{4}{1}{57--75}.
\PrintBackRefs{\CurrentBib}

\bibitem [\protect \citeauthoryear {%
Mohammad%
, Dunne%
\BCBL {}\ \BBA {} Dorr%
}{%
Mohammad%
\ \protect \BOthers {.}}{%
{\protect \APACyear {2009}}%
}]{%
Mohammad2009}
\APACinsertmetastar {%
Mohammad2009}%
\begin{APACrefauthors}%
Mohammad, S.%
, Dunne, C.%
\BCBL {}\ \BBA {} Dorr, B.%
\end{APACrefauthors}%
\unskip\
\newblock
\APACrefYearMonthDay{2009}{}{}.
\newblock
{\BBOQ}\APACrefatitle {{Generating High-coverage Semantic Orientation Lexicons
  from Overtly Marked Words and a Thesaurus}} {{Generating High-coverage
  Semantic Orientation Lexicons from Overtly Marked Words and a
  Thesaurus}}.{\BBCQ}
\newblock
\BIn{} \APACrefbtitle {{Proceedings of the 2009 Conference on Empirical Methods
  in Natural Language Processing: Volume 2}} {{Proceedings of the 2009
  Conference on Empirical Methods in Natural Language Processing: Volume 2}}\
  (\BPGS\ 599--608).
\newblock
\APACaddressPublisher{Stroudsburg, PA, USA}{Association for Computational
  Linguistics}.
\PrintBackRefs{\CurrentBib}

\bibitem [\protect \citeauthoryear {%
Nadali%
, Azmi~Murad%
, Sharef%
, Mustapha%
\BCBL {}\ \BBA {} Shojaee%
}{%
Nadali%
\ \protect \BOthers {.}}{%
{\protect \APACyear {2013}}%
}]{%
Nadali2013}
\APACinsertmetastar {%
Nadali2013}%
\begin{APACrefauthors}%
Nadali, S.%
, Azmi~Murad, M\BPBI A.%
, Sharef, N\BPBI M.%
, Mustapha, A.%
\BCBL {}\ \BBA {} Shojaee, S.%
\end{APACrefauthors}%
\unskip\
\newblock
\APACrefYearMonthDay{2013}{}{}.
\newblock
{\BBOQ}\APACrefatitle {{A review of cyberbullying detection: An overview}} {{A
  review of cyberbullying detection: An overview}}.{\BBCQ}
\newblock
\BIn{} \APACrefbtitle {{13th International Conference on Intellient Systems
  Design and Applications}} {{13th International Conference on Intellient
  Systems Design and Applications}}\ (\BPGS\ 325--330).
\newblock
\APACaddressPublisher{Salangor, Malaysia,}{}.
\newblock
\begin{APACrefDOI} \doi{10.1109/ISDA.2013.6920758} \end{APACrefDOI}
\PrintBackRefs{\CurrentBib}

\bibitem [\protect \citeauthoryear {%
Nahar%
, Al-Maskari%
, Li%
\BCBL {}\ \BBA {} Pang%
}{%
Nahar%
\ \protect \BOthers {.}}{%
{\protect \APACyear {2014}}%
}]{%
Nahar2014}
\APACinsertmetastar {%
Nahar2014}%
\begin{APACrefauthors}%
Nahar, V.%
, Al-Maskari, S.%
, Li, X.%
\BCBL {}\ \BBA {} Pang, C.%
\end{APACrefauthors}%
\unskip\
\newblock
\APACrefYearMonthDay{2014}{}{}.
\newblock
{\BBOQ}\APACrefatitle {{Semi-supervised Learning for Cyberbullying Detection in
  Social Networks}} {{Semi-supervised Learning for Cyberbullying Detection in
  Social Networks}}.{\BBCQ}
\newblock
\BIn{} \APACrefbtitle {{ADC.Databases Theory and Applications}} {{ADC.Databases
  Theory and Applications}}\ (\BPGS\ 160--171).
\PrintBackRefs{\CurrentBib}

\bibitem [\protect \citeauthoryear {%
Nansel%
\ \protect \BOthers {.}}{%
Nansel%
\ \protect \BOthers {.}}{%
{\protect \APACyear {2001}}%
}]{%
Nansel2001}
\APACinsertmetastar {%
Nansel2001}%
\begin{APACrefauthors}%
Nansel, T\BPBI R.%
, Overpeck, M.%
, Pilla, R\BPBI S.%
, Ruan, J\BPBI W.%
, Morton, B\BPBI S.%
\BCBL {}\ \BBA {} Scheidt, P.%
\end{APACrefauthors}%
\unskip\
\newblock
\APACrefYearMonthDay{2001}{}{}.
\newblock
{\BBOQ}\APACrefatitle {{Bullying behaviors among US youth: prevalence and
  association with psychosocial adjustment}} {{Bullying behaviors among US
  youth: prevalence and association with psychosocial adjustment}}.{\BBCQ}
\newblock
\APACjournalVolNumPages{{JAMA}}{285}{16}{2094--2100}.
\PrintBackRefs{\CurrentBib}

\bibitem [\protect \citeauthoryear {%
Nielsen%
}{%
Nielsen%
}{%
{\protect \APACyear {2011}}%
}]{%
Afinn2011}
\APACinsertmetastar {%
Afinn2011}%
\begin{APACrefauthors}%
Nielsen, F\BPBI {\AA}.%
\end{APACrefauthors}%
\unskip\
\newblock
\APACrefYearMonthDay{2011}{}{}.
\newblock
{\BBOQ}\APACrefatitle {{A New ANEW: Evaluation of a Word List for Sentiment
  Analysis in Microblogs}} {{A New ANEW: Evaluation of a Word List for
  Sentiment Analysis in Microblogs}}.{\BBCQ}
\newblock
\BIn{} M.~Rowe, M.~Stankovic, A\BHBI S.~Dadzie\BCBL {}\ \BBA {} M.~Hardey\
  (\BEDS), \APACrefbtitle {{Proceedings of the ESWC2011 Workshop on `Making
  Sense of Microposts': Big things come in small packages}} {{Proceedings of
  the ESWC2011 Workshop on `Making Sense of Microposts': Big things come in
  small packages}}\ (\BVOL~718, \BPGS\ 93--98).
\newblock
\APACaddressPublisher{}{CEUR-WS.org}.
\PrintBackRefs{\CurrentBib}

\bibitem [\protect \citeauthoryear {%
Olweus%
}{%
Olweus%
}{%
{\protect \APACyear {1993}}%
}]{%
Olweus1993}
\APACinsertmetastar {%
Olweus1993}%
\begin{APACrefauthors}%
Olweus, D.%
\end{APACrefauthors}%
\unskip\
\newblock
\APACrefYear{1993}.
\newblock
\APACrefbtitle {{Bullying at School: What We Know and What We Can Do}}
  {{Bullying at School: What We Know and What We Can Do}}\ (\PrintOrdinal{2nd}\
  \BEd).
\newblock
\APACaddressPublisher{}{Wiley}.
\PrintBackRefs{\CurrentBib}

\bibitem [\protect \citeauthoryear {%
O'Moore%
\ \BBA {} Kirkham%
}{%
O'Moore%
\ \BBA {} Kirkham%
}{%
{\protect \APACyear {2001}}%
}]{%
Moore2001}
\APACinsertmetastar {%
Moore2001}%
\begin{APACrefauthors}%
O'Moore, M.%
\BCBT {}\ \BBA {} Kirkham, C.%
\end{APACrefauthors}%
\unskip\
\newblock
\APACrefYearMonthDay{2001}{}{}.
\newblock
{\BBOQ}\APACrefatitle {{Self-esteem and its relationship to bullying
  behaviour}} {{Self-esteem and its relationship to bullying
  behaviour}}.{\BBCQ}
\newblock
\APACjournalVolNumPages{{Aggressive Behavior}}{27}{4}{269--283}.
\PrintBackRefs{\CurrentBib}

\bibitem [\protect \citeauthoryear {%
Online%
}{%
Online%
}{%
{\protect \APACyear {2014}}%
}]{%
EUKidsOnline2014}
\APACinsertmetastar {%
EUKidsOnline2014}%
\begin{APACrefauthors}%
Online, E\BPBI K.%
\end{APACrefauthors}%
\unskip\
\newblock
\APACrefYearMonthDay{2014}{}{}.
\newblock
\APACrefbtitle {{EU Kids Online: findings, methods, recommendations. EU Kids
  Online, LSE, London, UK. http://eprints.lse.ac.uk/60512/}.} {{EU Kids Online:
  findings, methods, recommendations. EU Kids Online, LSE, London, UK.
  http://eprints.lse.ac.uk/60512/}.}
\newblock
\APACaddressPublisher{}{London: EU Kids Online}.
\newblock
\begin{APACrefURL} \url{http://eprints.lse.ac.uk/60512/} \end{APACrefURL}
\PrintBackRefs{\CurrentBib}

\bibitem [\protect \citeauthoryear {%
O'Sullivan%
\ \BBA {} Flanagin%
}{%
O'Sullivan%
\ \BBA {} Flanagin%
}{%
{\protect \APACyear {2003}}%
}]{%
Osullivan2003}
\APACinsertmetastar {%
Osullivan2003}%
\begin{APACrefauthors}%
O'Sullivan, P\BPBI B.%
\BCBT {}\ \BBA {} Flanagin, A\BPBI J.%
\end{APACrefauthors}%
\unskip\
\newblock
\APACrefYearMonthDay{2003}{}{}.
\newblock
{\BBOQ}\APACrefatitle {{Reconceptualizing `flaming' and other problematic
  messages}} {{Reconceptualizing `flaming' and other problematic
  messages}}.{\BBCQ}
\newblock
\APACjournalVolNumPages{{New Media \& Society}}{5}{1}{69-94}.
\PrintBackRefs{\CurrentBib}

\bibitem [\protect \citeauthoryear {%
Pedregosa%
\ \protect \BOthers {.}}{%
Pedregosa%
\ \protect \BOthers {.}}{%
{\protect \APACyear {2011}}%
}]{%
Pedregosa2011}
\APACinsertmetastar {%
Pedregosa2011}%
\begin{APACrefauthors}%
Pedregosa, F.%
, Varoquaux, G.%
, Gramfort, A.%
, Michel, V.%
, Thirion, B.%
, Grisel, O.%
\BDBL {}Duchesnay, E.%
\end{APACrefauthors}%
\unskip\
\newblock
\APACrefYearMonthDay{2011}{}{}.
\newblock
{\BBOQ}\APACrefatitle {{Scikit-learn: Machine Learning in Python}}
  {{Scikit-learn: Machine Learning in Python}}.{\BBCQ}
\newblock
\APACjournalVolNumPages{{Journal of Machine Learning
  Research}}{12}{}{2825--2830}.
\PrintBackRefs{\CurrentBib}

\bibitem [\protect \citeauthoryear {%
Pennebaker%
, Francis%
\BCBL {}\ \BBA {} Booth%
}{%
Pennebaker%
\ \protect \BOthers {.}}{%
{\protect \APACyear {2001}}%
}]{%
Pennebaker2001}
\APACinsertmetastar {%
Pennebaker2001}%
\begin{APACrefauthors}%
Pennebaker, J\BPBI W.%
, Francis, M\BPBI E.%
\BCBL {}\ \BBA {} Booth, R\BPBI J.%
\end{APACrefauthors}%
\unskip\
\newblock
\APACrefYear{2001}.
\newblock
\APACrefbtitle {{Linguistic Inquiry and Word Count: LIWC 2001}} {{Linguistic
  Inquiry and Word Count: LIWC 2001}}.
\newblock
\APACaddressPublisher{Mahwah, NJ}{Lawrence Erlbaum Associates}.
\PrintBackRefs{\CurrentBib}

\bibitem [\protect \citeauthoryear {%
Price%
\ \BBA {} Dalgleish%
}{%
Price%
\ \BBA {} Dalgleish%
}{%
{\protect \APACyear {2010}}%
}]{%
Price2010}
\APACinsertmetastar {%
Price2010}%
\begin{APACrefauthors}%
Price, M.%
\BCBT {}\ \BBA {} Dalgleish, J.%
\end{APACrefauthors}%
\unskip\
\newblock
\APACrefYearMonthDay{2010}{}{}.
\newblock
{\BBOQ}\APACrefatitle {{Cyberbullying: Experiences, Impacts and Coping
  Strategies as Described by Australian Young People}} {{Cyberbullying:
  Experiences, Impacts and Coping Strategies as Described by Australian Young
  People}}.{\BBCQ}
\newblock
\APACjournalVolNumPages{Youth Studies Australia}{29}{2}{51--59}.
\PrintBackRefs{\CurrentBib}

\bibitem [\protect \citeauthoryear {%
Rehurek%
\ \BBA {} Sojka%
}{%
Rehurek%
\ \BBA {} Sojka%
}{%
{\protect \APACyear {2010}}%
}]{%
Rehurek2010}
\APACinsertmetastar {%
Rehurek2010}%
\begin{APACrefauthors}%
Rehurek, R.%
\BCBT {}\ \BBA {} Sojka, P.%
\end{APACrefauthors}%
\unskip\
\newblock
\APACrefYearMonthDay{2010}{}{}.
\newblock
{\BBOQ}\APACrefatitle {{Software framework for topic modelling with large
  corpora}} {{Software framework for topic modelling with large
  corpora}}.{\BBCQ}
\newblock
\BIn{} \APACrefbtitle {{The LREC 2010 Workshop on new Challenges for NLP
  Frameworks}} {{The LREC 2010 Workshop on new Challenges for NLP Frameworks}}\
  (\BPGS\ 45--50).
\PrintBackRefs{\CurrentBib}

\bibitem [\protect \citeauthoryear {%
Reynolds%
, Kontostathis%
\BCBL {}\ \BBA {} Edwards%
}{%
Reynolds%
\ \protect \BOthers {.}}{%
{\protect \APACyear {2011}}%
}]{%
Reynolds2011}
\APACinsertmetastar {%
Reynolds2011}%
\begin{APACrefauthors}%
Reynolds, K.%
, Kontostathis, A.%
\BCBL {}\ \BBA {} Edwards, L.%
\end{APACrefauthors}%
\unskip\
\newblock
\APACrefYearMonthDay{2011}{}{}.
\newblock
{\BBOQ}\APACrefatitle {{Using Machine Learning to Detect Cyberbullying}}
  {{Using Machine Learning to Detect Cyberbullying}}.{\BBCQ}
\newblock
\BIn{} \APACrefbtitle {{Proceedings of the 2011 10th International Conference
  on Machine Learning and Applications and Workshops}} {{Proceedings of the
  2011 10th International Conference on Machine Learning and Applications and
  Workshops}}\ (\BPGS\ 241--244).
\newblock
\APACaddressPublisher{Washington, DC, USA}{{IEEE Computer Society}}.
\PrintBackRefs{\CurrentBib}

\bibitem [\protect \citeauthoryear {%
Royen%
, Poels%
\BCBL {}\ \BBA {} Vandebosch%
}{%
Royen%
\ \protect \BOthers {.}}{%
{\protect \APACyear {2016}}%
}]{%
VanRoyen2016}
\APACinsertmetastar {%
VanRoyen2016}%
\begin{APACrefauthors}%
Royen, K\BPBI V.%
, Poels, K.%
\BCBL {}\ \BBA {} Vandebosch, H.%
\end{APACrefauthors}%
\unskip\
\newblock
\APACrefYearMonthDay{2016}{}{}.
\newblock
{\BBOQ}\APACrefatitle {{Harmonizing freedom and protection: Adolescents' voices
  on automatic monitoring of social networking sites}} {{Harmonizing freedom
  and protection: Adolescents' voices on automatic monitoring of social
  networking sites}}.{\BBCQ}
\newblock
\APACjournalVolNumPages{{Children and Youth Services Review}}{64}{}{35 - 41}.
\newblock
\begin{APACrefDOI} \doi{http://dx.doi.org/10.1016/j.childyouth.2016.02.024}
  \end{APACrefDOI}
\PrintBackRefs{\CurrentBib}

\bibitem [\protect \citeauthoryear {%
Salmivalli%
}{%
Salmivalli%
}{%
{\protect \APACyear {2010}}%
}]{%
Salmivalli2010}
\APACinsertmetastar {%
Salmivalli2010}%
\begin{APACrefauthors}%
Salmivalli, C.%
\end{APACrefauthors}%
\unskip\
\newblock
\APACrefYearMonthDay{2010}{}{}.
\newblock
{\BBOQ}\APACrefatitle {{Bullying and the peer group: A review}} {{Bullying and
  the peer group: A review}}.{\BBCQ}
\newblock
\APACjournalVolNumPages{{Aggression and Violent Behavior}}{15}{2}{112--120}.
\PrintBackRefs{\CurrentBib}

\bibitem [\protect \citeauthoryear {%
Salmivalli%
, Kaukiainen%
, Kaistaniemi%
\BCBL {}\ \BBA {} Lagerspetz%
}{%
Salmivalli%
\ \protect \BOthers {.}}{%
{\protect \APACyear {1999}}%
}]{%
Salmivalli1999}
\APACinsertmetastar {%
Salmivalli1999}%
\begin{APACrefauthors}%
Salmivalli, C.%
, Kaukiainen, A.%
, Kaistaniemi, L.%
\BCBL {}\ \BBA {} Lagerspetz, K\BPBI M\BPBI J.%
\end{APACrefauthors}%
\unskip\
\newblock
\APACrefYearMonthDay{1999}{}{}.
\newblock
{\BBOQ}\APACrefatitle {{Self-Evaluated Self-Esteem, Peer-Evaluated Self-Esteem,
  and Defensive Egotism as Predictors of Adolescents' Participation in Bullying
  Situations}} {{Self-Evaluated Self-Esteem, Peer-Evaluated Self-Esteem, and
  Defensive Egotism as Predictors of Adolescents' Participation in Bullying
  Situations}}.{\BBCQ}
\newblock
\APACjournalVolNumPages{{Personality and Social Psychology
  Bulletin}}{25}{10}{1268--1278}.
\newblock
\begin{APACrefDOI} \doi{10.1177/0146167299258008} \end{APACrefDOI}
\PrintBackRefs{\CurrentBib}

\bibitem [\protect \citeauthoryear {%
Salmivalli%
, Lagerspetz%
, Bj\"{o}rkqvist%
, \"{O}sterman%
\BCBL {}\ \BBA {} Kaukiainen%
}{%
Salmivalli%
\ \protect \BOthers {.}}{%
{\protect \APACyear {1996}}%
}]{%
Salmivalli1996}
\APACinsertmetastar {%
Salmivalli1996}%
\begin{APACrefauthors}%
Salmivalli, C.%
, Lagerspetz, K.%
, Bj\"{o}rkqvist, K.%
, \"{O}sterman, K.%
\BCBL {}\ \BBA {} Kaukiainen, A.%
\end{APACrefauthors}%
\unskip\
\newblock
\APACrefYearMonthDay{1996}{}{}.
\newblock
{\BBOQ}\APACrefatitle {{Bullying as a group process: Participant roles and
  their relations to social status within the group}} {{Bullying as a group
  process: Participant roles and their relations to social status within the
  group}}.{\BBCQ}
\newblock
\APACjournalVolNumPages{{Aggressive Behavior}}{22}{1}{1--15}.
\PrintBackRefs{\CurrentBib}

\bibitem [\protect \citeauthoryear {%
Salmivalli%
, Voeten%
\BCBL {}\ \BBA {} Poskiparta%
}{%
Salmivalli%
\ \protect \BOthers {.}}{%
{\protect \APACyear {2011}}%
}]{%
Salmivalli2011}
\APACinsertmetastar {%
Salmivalli2011}%
\begin{APACrefauthors}%
Salmivalli, C.%
, Voeten, M.%
\BCBL {}\ \BBA {} Poskiparta, E.%
\end{APACrefauthors}%
\unskip\
\newblock
\APACrefYearMonthDay{2011}{}{}.
\newblock
{\BBOQ}\APACrefatitle {{Bystanders Matter: Associations Between Reinforcing,
  Defending, and the Frequency of Bullying Behavior in Classrooms}}
  {{Bystanders Matter: Associations Between Reinforcing, Defending, and the
  Frequency of Bullying Behavior in Classrooms}}.{\BBCQ}
\newblock
\APACjournalVolNumPages{{Journal of Clinical Child \& Adolescent
  Psychology}}{40}{5}{668-676}.
\newblock
\begin{APACrefDOI} \doi{10.1080/15374416.2011.597090} \end{APACrefDOI}
\PrintBackRefs{\CurrentBib}

\bibitem [\protect \citeauthoryear {%
Slonje%
\ \BBA {} Smith%
}{%
Slonje%
\ \BBA {} Smith%
}{%
{\protect \APACyear {2008}}%
}]{%
Slonje2008}
\APACinsertmetastar {%
Slonje2008}%
\begin{APACrefauthors}%
Slonje, R.%
\BCBT {}\ \BBA {} Smith, P\BPBI K.%
\end{APACrefauthors}%
\unskip\
\newblock
\APACrefYearMonthDay{2008}{}{}.
\newblock
{\BBOQ}\APACrefatitle {{Cyberbullying: Another main type of bullying?}}
  {{Cyberbullying: Another main type of bullying?}}{\BBCQ}
\newblock
\APACjournalVolNumPages{{Scandinavian Journal of Psychology}}{49}{2}{147--154}.
\PrintBackRefs{\CurrentBib}

\bibitem [\protect \citeauthoryear {%
Slonje%
, Smith%
\BCBL {}\ \BBA {} Fris\'{e}n%
}{%
Slonje%
\ \protect \BOthers {.}}{%
{\protect \APACyear {2013}}%
}]{%
Slonje2013}
\APACinsertmetastar {%
Slonje2013}%
\begin{APACrefauthors}%
Slonje, R.%
, Smith, P\BPBI K.%
\BCBL {}\ \BBA {} Fris\'{e}n, A.%
\end{APACrefauthors}%
\unskip\
\newblock
\APACrefYearMonthDay{2013}{}{}.
\newblock
{\BBOQ}\APACrefatitle {{The Nature of Cyberbullying, and Strategies for
  Prevention}} {{The Nature of Cyberbullying, and Strategies for
  Prevention}}.{\BBCQ}
\newblock
\APACjournalVolNumPages{{Compututers in Human Behavior}}{29}{1}{26--32}.
\PrintBackRefs{\CurrentBib}

\bibitem [\protect \citeauthoryear {%
Smith%
\ \protect \BOthers {.}}{%
Smith%
\ \protect \BOthers {.}}{%
{\protect \APACyear {2008}}%
}]{%
Smith2008}
\APACinsertmetastar {%
Smith2008}%
\begin{APACrefauthors}%
Smith, P\BPBI K.%
, Mahdavi, J.%
, Carvalho, M.%
, Fisher, S.%
, Russell, S.%
\BCBL {}\ \BBA {} Tippett, N.%
\end{APACrefauthors}%
\unskip\
\newblock
\APACrefYearMonthDay{2008}{}{}.
\newblock
{\BBOQ}\APACrefatitle {{Cyberbullying: its nature and impact in secondary
  school pupils}} {{Cyberbullying: its nature and impact in secondary school
  pupils}}.{\BBCQ}
\newblock
\APACjournalVolNumPages{Journal of Child Psychology and
  Psychiatry}{49}{4}{376--385}.
\PrintBackRefs{\CurrentBib}

\bibitem [\protect \citeauthoryear {%
Stenetorp%
\ \protect \BOthers {.}}{%
Stenetorp%
\ \protect \BOthers {.}}{%
{\protect \APACyear {2012}}%
}]{%
Stenetorp2012}
\APACinsertmetastar {%
Stenetorp2012}%
\begin{APACrefauthors}%
Stenetorp, P.%
, Pyysalo, S.%
, Topi\'{c}, G.%
, Ohta, T.%
, Ananiadou, S.%
\BCBL {}\ \BBA {} Tsujii, J.%
\end{APACrefauthors}%
\unskip\
\newblock
\APACrefYearMonthDay{2012}{}{}.
\newblock
{\BBOQ}\APACrefatitle {{brat: a Web-based Tool for NLP-Assisted Text
  Annotation}} {{brat: a Web-based Tool for NLP-Assisted Text
  Annotation}}.{\BBCQ}
\newblock
\BIn{} \APACrefbtitle {{Proceedings of the Demonstrations Session at EACL
  2012}} {{Proceedings of the Demonstrations Session at EACL 2012}}\ (\BPGS\
  102--107).
\newblock
\APACaddressPublisher{Avignon, France}{}.
\PrintBackRefs{\CurrentBib}

\bibitem [\protect \citeauthoryear {%
Stone%
, Dunphy%
, Smith%
\BCBL {}\ \BBA {} Ogilvie%
}{%
Stone%
\ \protect \BOthers {.}}{%
{\protect \APACyear {1966}}%
}]{%
Stone1966}
\APACinsertmetastar {%
Stone1966}%
\begin{APACrefauthors}%
Stone, P\BPBI J.%
, Dunphy, D\BPBI C\BPBI D.%
, Smith, M\BPBI S.%
\BCBL {}\ \BBA {} Ogilvie, D\BPBI M.%
\end{APACrefauthors}%
\unskip\
\newblock
\APACrefYear{1966}.
\newblock
\APACrefbtitle {{The General Inquirer: A Computer Approach to Content
  Analysis}} {{The General Inquirer: A Computer Approach to Content Analysis}}.
\newblock
\APACaddressPublisher{}{The MIT Press}.
\PrintBackRefs{\CurrentBib}

\bibitem [\protect \citeauthoryear {%
Sui%
}{%
Sui%
}{%
{\protect \APACyear {2015}}%
}]{%
Sui2015}
\APACinsertmetastar {%
Sui2015}%
\begin{APACrefauthors}%
Sui, J.%
\end{APACrefauthors}%
\unskip\
\newblock
\APACrefYear{2015}.
\unskip\
\newblock
\APACrefbtitle {{Understanding and Fighting Bullying with Machine Learning}}
  {{Understanding and Fighting Bullying with Machine Learning}}\
  \APACtypeAddressSchool {PhD thesis}{}{}.
\unskip\
\newblock
\APACaddressSchool {}{Department of Computer Sciences, University of
  Wisconsin-Madison}.
\PrintBackRefs{\CurrentBib}

\bibitem [\protect \citeauthoryear {%
Tokunaga%
}{%
Tokunaga%
}{%
{\protect \APACyear {2010}}%
}]{%
Tokunaga2010}
\APACinsertmetastar {%
Tokunaga2010}%
\begin{APACrefauthors}%
Tokunaga, R\BPBI S.%
\end{APACrefauthors}%
\unskip\
\newblock
\APACrefYearMonthDay{2010}{}{}.
\newblock
{\BBOQ}\APACrefatitle {{Following You Home from School: A Critical Review and
  Synthesis of Research on Cyberbullying Victimization}} {{Following You Home
  from School: A Critical Review and Synthesis of Research on Cyberbullying
  Victimization}}.{\BBCQ}
\newblock
\APACjournalVolNumPages{{Computers in Human Behavior}}{26}{3}{277--287}.
\PrintBackRefs{\CurrentBib}

\bibitem [\protect \citeauthoryear {%
Van~Cleemput%
\ \protect \BOthers {.}}{%
Van~Cleemput%
\ \protect \BOthers {.}}{%
{\protect \APACyear {2013}}%
}]{%
VanCleemput2013}
\APACinsertmetastar {%
VanCleemput2013}%
\begin{APACrefauthors}%
Van~Cleemput, K.%
, Bastiaensens, S.%
, Vandebosch, H.%
, Poels, K.%
, Deboutte, G.%
, DeSmet, A.%
\BCBL {}\ \BBA {} De~Bourdeaudhuij, I.%
\end{APACrefauthors}%
\unskip\
\newblock
\APACrefYearMonthDay{2013}{}{}.
\newblock
\APACrefbtitle {{Zes jaar onderzoek naar cyberpesten in Vlaanderen, Belgi\"{e}
  en daarbuiten: een overzicht van de bevindingen. (Six years of research on
  cyberbullying in Flanders, Belgium and beyond: an overview of the findings.)
  (White Paper)}} {{Zes jaar onderzoek naar cyberpesten in Vlaanderen,
  Belgi\"{e} en daarbuiten: een overzicht van de bevindingen. (Six years of
  research on cyberbullying in Flanders, Belgium and beyond: an overview of the
  findings.) (White Paper)}}\ \APACbVolEdTR{}{\BTR{}}.
\newblock
\APACaddressInstitution{}{University of Antwerp \& Ghent University}.
\PrintBackRefs{\CurrentBib}

\bibitem [\protect \citeauthoryear {%
Vandebosch%
\ \BBA {} Van~Cleemput%
}{%
Vandebosch%
\ \BBA {} Van~Cleemput%
}{%
{\protect \APACyear {2008}}%
}]{%
Vandebosch2008}
\APACinsertmetastar {%
Vandebosch2008}%
\begin{APACrefauthors}%
Vandebosch, H.%
\BCBT {}\ \BBA {} Van~Cleemput, K.%
\end{APACrefauthors}%
\unskip\
\newblock
\APACrefYearMonthDay{2008}{}{}.
\newblock
{\BBOQ}\APACrefatitle {{Defining cyberbullying: a qualitative research into the
  perceptions of youngsters}} {{Defining cyberbullying: a qualitative research
  into the perceptions of youngsters}}.{\BBCQ}
\newblock
\APACjournalVolNumPages{{Cyberpsychology and behavior: the impact of the
  Internet, multimedia and virtual reality on behavior and
  society}}{11}{4}{499--503}.
\PrintBackRefs{\CurrentBib}

\bibitem [\protect \citeauthoryear {%
Vandebosch%
\ \BBA {} Van~Cleemput%
}{%
Vandebosch%
\ \BBA {} Van~Cleemput%
}{%
{\protect \APACyear {2009}}%
}]{%
Vandebosch2009}
\APACinsertmetastar {%
Vandebosch2009}%
\begin{APACrefauthors}%
Vandebosch, H.%
\BCBT {}\ \BBA {} Van~Cleemput, K.%
\end{APACrefauthors}%
\unskip\
\newblock
\APACrefYearMonthDay{2009}{}{}.
\newblock
{\BBOQ}\APACrefatitle {{Cyberbullying among youngsters: profiles of bullies and
  victims.}} {{Cyberbullying among youngsters: profiles of bullies and
  victims.}}{\BBCQ}
\newblock
\APACjournalVolNumPages{{New Media \& Society}}{11}{8}{1349--1371}.
\PrintBackRefs{\CurrentBib}

\bibitem [\protect \citeauthoryear {%
Vandebosch%
, Van~Cleemput%
, Mortelmans%
\BCBL {}\ \BBA {} Walrave%
}{%
Vandebosch%
\ \protect \BOthers {.}}{%
{\protect \APACyear {2006}}%
}]{%
Vandebosch2006}
\APACinsertmetastar {%
Vandebosch2006}%
\begin{APACrefauthors}%
Vandebosch, H.%
, Van~Cleemput, K.%
, Mortelmans, D.%
\BCBL {}\ \BBA {} Walrave, M.%
\end{APACrefauthors}%
\unskip\
\newblock
\APACrefYearMonthDay{2006}{}{}.
\newblock
\APACrefbtitle {{Cyberpesten bij jongeren in Vlaanderen: Een studie in opdracht
  van het viWTA (Cyberbullying among youngsters in Flanders: a study
  commissoned by the viWTA). Brussels: viWTA}} {{Cyberpesten bij jongeren in
  Vlaanderen: Een studie in opdracht van het viWTA (Cyberbullying among
  youngsters in Flanders: a study commissoned by the viWTA). Brussels: viWTA}}\
  \APACbVolEdTR{}{\BTR{}}.
\newblock
\begin{APACrefURL}
  \url{http://www.samenlevingentechnologie.be/ists/nl/pdf/rapporten\\/rapportcyberpesten.pdf}
  \end{APACrefURL}
\PrintBackRefs{\CurrentBib}

\bibitem [\protect \citeauthoryear {%
van~de Kauter%
\ \protect \BOthers {.}}{%
van~de Kauter%
\ \protect \BOthers {.}}{%
{\protect \APACyear {2013}}%
}]{%
vandeKauter2013}
\APACinsertmetastar {%
vandeKauter2013}%
\begin{APACrefauthors}%
van~de Kauter, M.%
, Coorman, G.%
, Lefever, E.%
, Desmet, B.%
, Macken, L.%
\BCBL {}\ \BBA {} Hoste, V.%
\end{APACrefauthors}%
\unskip\
\newblock
\APACrefYearMonthDay{2013}{}{}.
\newblock
{\BBOQ}\APACrefatitle {{LeTs Preprocess: The multilingual LT3 linguistic
  preprocessing toolkit}} {{LeTs Preprocess: The multilingual LT3 linguistic
  preprocessing toolkit}}.{\BBCQ}
\newblock
\APACjournalVolNumPages{{Computational Linguistics in the Netherlands
  Journal}}{3}{}{103--120}.
\PrintBackRefs{\CurrentBib}

\bibitem [\protect \citeauthoryear {%
Van~Hee%
, Lefever%
\BCBL {}\ \protect \BOthers {.}}{%
Van~Hee%
, Lefever%
\BCBL {}\ \protect \BOthers {.}}{%
{\protect \APACyear {2015}}%
{\protect \APACexlab {{\protect \BCnt {1}}}}}]{%
VanHee2015b}
\APACinsertmetastar {%
VanHee2015b}%
\begin{APACrefauthors}%
Van~Hee, C.%
, Lefever, E.%
, Verhoeven, B.%
, Mennes, J.%
, Desmet, B.%
, De~Pauw, G.%
\BDBL {}Hoste, V.%
\end{APACrefauthors}%
\unskip\
\newblock
\APACrefYearMonthDay{2015{\protect \BCnt {1}}}{}{}.
\newblock
{\BBOQ}\APACrefatitle {{Automatic detection and prevention of cyberbullying}}
  {{Automatic detection and prevention of cyberbullying}}.{\BBCQ}
\newblock
\BIn{} P.~Lorenz\ \BBA {} C.~Bourret\ (\BEDS), \APACrefbtitle {International
  Conference on Human and Social Analytics, Proceedings} {International
  conference on human and social analytics, proceedings}\ (\BPGS\ 13--18).
\newblock
\APACaddressPublisher{}{IARIA}.
\PrintBackRefs{\CurrentBib}

\bibitem [\protect \citeauthoryear {%
Van~Hee%
, Lefever%
\BCBL {}\ \protect \BOthers {.}}{%
Van~Hee%
, Lefever%
\BCBL {}\ \protect \BOthers {.}}{%
{\protect \APACyear {2015}}%
{\protect \APACexlab {{\protect \BCnt {2}}}}}]{%
VanHee2015a}
\APACinsertmetastar {%
VanHee2015a}%
\begin{APACrefauthors}%
Van~Hee, C.%
, Lefever, E.%
, Verhoeven, B.%
, Mennes, J.%
, Desmet, B.%
, De~Pauw, G.%
\BDBL {}Hoste, V.%
\end{APACrefauthors}%
\unskip\
\newblock
\APACrefYearMonthDay{2015{\protect \BCnt {2}}}{}{}.
\newblock
{\BBOQ}\APACrefatitle {{Detection and fine-grained classification of
  cyberbullying events}} {{Detection and fine-grained classification of
  cyberbullying events}}.{\BBCQ}
\newblock
\BIn{} G.~Angelova, K.~Bontcheva\BCBL {}\ \BBA {} R.~Mitkov\ (\BEDS),
  \APACrefbtitle {Proceedings of Recent Advances in Natural Language
  Processing, Proceedings} {Proceedings of recent advances in natural language
  processing, proceedings}\ (\BPGS\ 672--680).
\PrintBackRefs{\CurrentBib}

\bibitem [\protect \citeauthoryear {%
Van~Hee%
, Verhoeven%
\BCBL {}\ \protect \BOthers {.}}{%
Van~Hee%
, Verhoeven%
\BCBL {}\ \protect \BOthers {.}}{%
{\protect \APACyear {2015}}%
}]{%
VanHee2015c}
\APACinsertmetastar {%
VanHee2015c}%
\begin{APACrefauthors}%
Van~Hee, C.%
, Verhoeven, B.%
, Lefever, E.%
, De~Pauw, G.%
, Daelemans, W.%
\BCBL {}\ \BBA {} Hoste, V.%
\end{APACrefauthors}%
\unskip\
\newblock
\APACrefYearMonthDay{2015}{}{}.
\newblock
\APACrefbtitle {{Guidelines for the Fine-Grained Analysis of Cyberbullying,
  version 1.0}} {{Guidelines for the Fine-Grained Analysis of Cyberbullying,
  version 1.0}}\ \APACbVolEdTR{}{\BTR{}\ \BNUM\ LT3 15-01}.
\newblock
\APACaddressInstitution{}{LT3, Language and Translation Technology Team--Ghent
  University}.
\PrintBackRefs{\CurrentBib}

\bibitem [\protect \citeauthoryear {%
Van~Royen%
, Poels%
, Daelemans%
\BCBL {}\ \BBA {} Vandebosch%
}{%
Van~Royen%
\ \protect \BOthers {.}}{%
{\protect \APACyear {2014}}%
}]{%
VanRoyen2014}
\APACinsertmetastar {%
VanRoyen2014}%
\begin{APACrefauthors}%
Van~Royen, K.%
, Poels, K.%
, Daelemans, W.%
\BCBL {}\ \BBA {} Vandebosch, H.%
\end{APACrefauthors}%
\unskip\
\newblock
\APACrefYearMonthDay{2014}{}{}.
\newblock
{\BBOQ}\APACrefatitle {{Automatic monitoring of cyberbullying on social
  networking sites: From technological feasibility to desirability}}
  {{Automatic monitoring of cyberbullying on social networking sites: From
  technological feasibility to desirability}}.{\BBCQ}
\newblock
\APACjournalVolNumPages{{Telematics and Informatics}}{}{}{}.
\newblock
\begin{APACrefDOI} \doi{10.1016/j.tele.2014.04.002} \end{APACrefDOI}
\PrintBackRefs{\CurrentBib}

\bibitem [\protect \citeauthoryear {%
Willard%
}{%
Willard%
}{%
{\protect \APACyear {2007}}%
}]{%
Willard2007}
\APACinsertmetastar {%
Willard2007}%
\begin{APACrefauthors}%
Willard, N\BPBI E.%
\end{APACrefauthors}%
\unskip\
\newblock
\APACrefYear{2007}.
\newblock
\APACrefbtitle {{Cyberbullying and Cyberthreats: Responding to the Challenge of
  Online Social Aggression, Threats, and Distress}} {{Cyberbullying and
  Cyberthreats: Responding to the Challenge of Online Social Aggression,
  Threats, and Distress}}\ (\PrintOrdinal{2nd}\ \BEd).
\newblock
\APACaddressPublisher{}{Research Publishers LLC}.
\PrintBackRefs{\CurrentBib}

\bibitem [\protect \citeauthoryear {%
Wilson%
, Wiebe%
\BCBL {}\ \BBA {} Hoffmann%
}{%
Wilson%
\ \protect \BOthers {.}}{%
{\protect \APACyear {2005}}%
}]{%
Wilson2005}
\APACinsertmetastar {%
Wilson2005}%
\begin{APACrefauthors}%
Wilson, T.%
, Wiebe, J.%
\BCBL {}\ \BBA {} Hoffmann, P.%
\end{APACrefauthors}%
\unskip\
\newblock
\APACrefYearMonthDay{2005}{}{}.
\newblock
{\BBOQ}\APACrefatitle {{Recognizing Contextual Polarity in Phrase-level
  Sentiment Analysis}} {{Recognizing Contextual Polarity in Phrase-level
  Sentiment Analysis}}.{\BBCQ}
\newblock
\BIn{} \APACrefbtitle {{Proceedings of the Conference on Human Language
  Technology and Empirical Methods in Natural Language Processing}}
  {{Proceedings of the Conference on Human Language Technology and Empirical
  Methods in Natural Language Processing}}\ (\BPGS\ 347--354).
\newblock
\APACaddressPublisher{}{Association for Computational Linguistics}.
\newblock
\begin{APACrefDOI} \doi{10.3115/1220575.1220619} \end{APACrefDOI}
\PrintBackRefs{\CurrentBib}

\bibitem [\protect \citeauthoryear {%
Wolak%
, Mitchell%
\BCBL {}\ \BBA {} Finkelhor%
}{%
Wolak%
\ \protect \BOthers {.}}{%
{\protect \APACyear {2007}}%
}]{%
Wolak2007}
\APACinsertmetastar {%
Wolak2007}%
\begin{APACrefauthors}%
Wolak, J.%
, Mitchell, K\BPBI J.%
\BCBL {}\ \BBA {} Finkelhor, D.%
\end{APACrefauthors}%
\unskip\
\newblock
\APACrefYearMonthDay{2007}{}{}.
\newblock
{\BBOQ}\APACrefatitle {{Does Online Harassment Constitute Bullying? An
  Exploration of Online Harassment by Known Peers and Online-Only Contacts}}
  {{Does Online Harassment Constitute Bullying? An Exploration of Online
  Harassment by Known Peers and Online-Only Contacts}}.{\BBCQ}
\newblock
\APACjournalVolNumPages{{Journal of Adolescent Health}}{41}{6,
  Supplement}{S51--S58}.
\PrintBackRefs{\CurrentBib}

\bibitem [\protect \citeauthoryear {%
Xu%
, Jun%
, Zhu%
\BCBL {}\ \BBA {} Bellmore%
}{%
Xu%
\ \protect \BOthers {.}}{%
{\protect \APACyear {2012}}%
}]{%
Xu2012}
\APACinsertmetastar {%
Xu2012}%
\begin{APACrefauthors}%
Xu, J\BHBI M.%
, Jun, K\BHBI S.%
, Zhu, X.%
\BCBL {}\ \BBA {} Bellmore, A.%
\end{APACrefauthors}%
\unskip\
\newblock
\APACrefYearMonthDay{2012}{}{}.
\newblock
{\BBOQ}\APACrefatitle {{Learning from Bullying Traces in Social Media}}
  {{Learning from Bullying Traces in Social Media}}.{\BBCQ}
\newblock
\BIn{} \APACrefbtitle {{Proceedings of the 2012 Conference of the North
  American Chapter of the Association for Computational Linguistics: Human
  Language Technologies}} {{Proceedings of the 2012 Conference of the North
  American Chapter of the Association for Computational Linguistics: Human
  Language Technologies}}\ (\BPGS\ 656--666).
\newblock
\APACaddressPublisher{Stroudsburg, PA, USA}{Association for Computational
  Linguistics}.
\PrintBackRefs{\CurrentBib}

\bibitem [\protect \citeauthoryear {%
Yin%
\ \protect \BOthers {.}}{%
Yin%
\ \protect \BOthers {.}}{%
{\protect \APACyear {2009}}%
}]{%
Yin2009}
\APACinsertmetastar {%
Yin2009}%
\begin{APACrefauthors}%
Yin, D.%
, Davison, B\BPBI D.%
, Xue, Z.%
, Hong, L.%
, Kontostathis, A.%
\BCBL {}\ \BBA {} Edwards, L.%
\end{APACrefauthors}%
\unskip\
\newblock
\APACrefYearMonthDay{2009}{}{}.
\newblock
{\BBOQ}\APACrefatitle {{Detection of Harassment on Web 2.0}} {{Detection of
  Harassment on Web 2.0}}.{\BBCQ}
\newblock
\BIn{} \APACrefbtitle {{Proceedings of the Content Analysis in the Web 2.0
  (CAW2.0)}.} {{Proceedings of the Content Analysis in the Web 2.0 (CAW2.0)}.}
\newblock
\APACaddressPublisher{Madrid, Spain}{}.
\PrintBackRefs{\CurrentBib}

\bibitem [\protect \citeauthoryear {%
Zhao%
, Zhou%
\BCBL {}\ \BBA {} Mao%
}{%
Zhao%
\ \protect \BOthers {.}}{%
{\protect \APACyear {2016}}%
}]{%
Zhao2016}
\APACinsertmetastar {%
Zhao2016}%
\begin{APACrefauthors}%
Zhao, R.%
, Zhou, A.%
\BCBL {}\ \BBA {} Mao, K.%
\end{APACrefauthors}%
\unskip\
\newblock
\APACrefYearMonthDay{2016}{}{}.
\newblock
{\BBOQ}\APACrefatitle {{Automatic Detection of Cyberbullying on Social Networks
  Based on Bullying Features}} {{Automatic Detection of Cyberbullying on Social
  Networks Based on Bullying Features}}.{\BBCQ}
\newblock
\BIn{} \APACrefbtitle {{Proceedings of the 17th International Conference on
  Distributed Computing and Networking}} {{Proceedings of the 17th
  International Conference on Distributed Computing and Networking}}\ (\BPGS\
  43:1--43:6).
\newblock
\APACaddressPublisher{New York, NY, USA}{ACM}.
\newblock
\begin{APACrefDOI} \doi{10.1145/2833312.2849567} \end{APACrefDOI}
\PrintBackRefs{\CurrentBib}

\bibitem [\protect \citeauthoryear {%
Zijlstra%
, Van~Meerveld%
, Van~Middendorp%
, Pennebaker%
\BCBL {}\ \BBA {} Geenen%
}{%
Zijlstra%
\ \protect \BOthers {.}}{%
{\protect \APACyear {2004}}%
}]{%
Zijlstra2004}
\APACinsertmetastar {%
Zijlstra2004}%
\begin{APACrefauthors}%
Zijlstra, H.%
, Van~Meerveld, T.%
, Van~Middendorp, H.%
, Pennebaker, J\BPBI W.%
\BCBL {}\ \BBA {} Geenen, R.%
\end{APACrefauthors}%
\unskip\
\newblock
\APACrefYearMonthDay{2004}{}{}.
\newblock
{\BBOQ}\APACrefatitle {{De Nederlandse versie van de `linguistic inquiry and
  word count' (LIWC)}} {{De Nederlandse versie van de `linguistic inquiry and
  word count' (LIWC)}}.{\BBCQ}
\newblock
\APACjournalVolNumPages{{Gedrag Gezond}}{32}{}{271--281}.
\PrintBackRefs{\CurrentBib}

\end{thebibliography}

\end{document}